\documentclass[11pt]{article}

\usepackage[preprint]{acl}

\usepackage{fontspec}
\defaultfontfeatures{Ligatures=TeX}
\usepackage{url}
\usepackage{placeins}
\usepackage{amsmath}
\usepackage{amssymb}
\usepackage{ragged2e}   % 智能左对齐，解决双栏词距问题
\justifying             % 全局智能排版，替代暴力两端对齐
\usepackage{booktabs}
\usepackage{graphicx}    % 插入图片
\usepackage{subcaption}  % 子图排版（替代旧的subfigure，更稳定）
\usepackage{listings}
\usepackage{tikz-cd}
\usepackage{placeins}
\usepackage{multicol}
\usepackage{multirow}
\usepackage{float} % 用于强制固定浮动体位置
\usepackage[table]{xcolor}
\usepackage{amsthm} % 定理/命题/证明专用宏包


\usepackage{mathtools}
\usepackage{algorithm}
\usepackage{algpseudocode}
\makeatletter
\newenvironment{breakablealgorithm}
  {%
   \begin{center}
     \refstepcounter{algorithm}%
     \hrule height .8pt depth 0pt \kern 2pt%
     \renewcommand{\caption}[2][\relax]{%
       {\raggedright\textbf{Algorithm \thealgorithm} ##2\par}%
       \ifx\relax##1\relax
         \addcontentsline{loa}{algorithm}{\protect\numberline{\thealgorithm}##2}%
       \else
         \addcontentsline{loa}{algorithm}{\protect\numberline{\thealgorithm}##1}%
       \fi
       \kern 2pt\hrule\kern 2pt
     }%
  }
  {%
     \kern 2pt\hrule\relax%
   \end{center}
  }
\makeatother

\usepackage[most]{tcolorbox} % 加载tcolorbox全功能
\tcbuselibrary{skins}   % 启用增强样式（支持标题栏右对齐）
\newcommand{\mediumtoken}[1]{\colorbox{red!45}{#1}}  % 中红（中等权重）
\newcommand{\heavytoken}[1]{\colorbox{red!70}{#1}}   % 深红（高权重/核心token）

\title{PAEC: Position-Aware Entropy Calibration for LLM Reasoning in RLVR}
\author{
Shumeng Yang$^{1,2}$ \quad
Yisu Liu$^{3}$ \quad
Jiayi Zheng$^{4,5}$ \quad
Zhaohui Yang$^{1,2}$ \quad
Linjing Li$^{1}$\thanks{Corresponding author.} \\
$^{1}$Institute of Automation, Chinese Academy of Sciences \\
$^{2}$School of Artificial Intelligence, University of Chinese Academy of Sciences \\
$^{3}$School of Artificial Intelligence, Beijing University of Posts and Telecommunications \\
$^{4}$Institute of Computing Technology, Chinese Academy of Sciences \\
$^{5}$School of Computer Science and Technology, University of Chinese Academy of Sciences \\
\texttt{yangshumeng2025@ia.ac.cn}
}

\begin{document}
\maketitle\raggedbottom

\begin{abstract}
\noindent Reinforcement learning with verifiable rewards (RLVR) improves large language model reasoning but often suffers from rapid policy-entropy collapse, where the policy prematurely concentrates on narrow high-probability reasoning paths. 
While global entropy regularization can encourage exploration, uniformly increasing entropy across all token positions is inefficient for long reasoning trajectories, where many tokens are not decision-relevant. 
We propose Position-Aware Entropy Calibration (PAEC), a token-level entropy-management framework that constructs a soft mask from local top-$p$ entropy and top-two candidate competition, and applies an anchor-based lower-bound penalty to prevent selected-position entropy collapse. 
Experiments on five mathematical reasoning benchmarks show that PAEC improves macro-average majority-vote performance over strong RLVR baselines, with clear gains on AIME-style tasks. 
Our results suggest that entropy management in reasoning RL should be formulated as selective exploration allocation over decision-sensitive positions rather than uniform randomness injection.
\end{abstract}

\begin{figure*}[t]  % 位置控制：h此处 t顶部 b底部 p单独页
  \centering         % 关键：图片居中
  \includegraphics[width=1.0\linewidth]{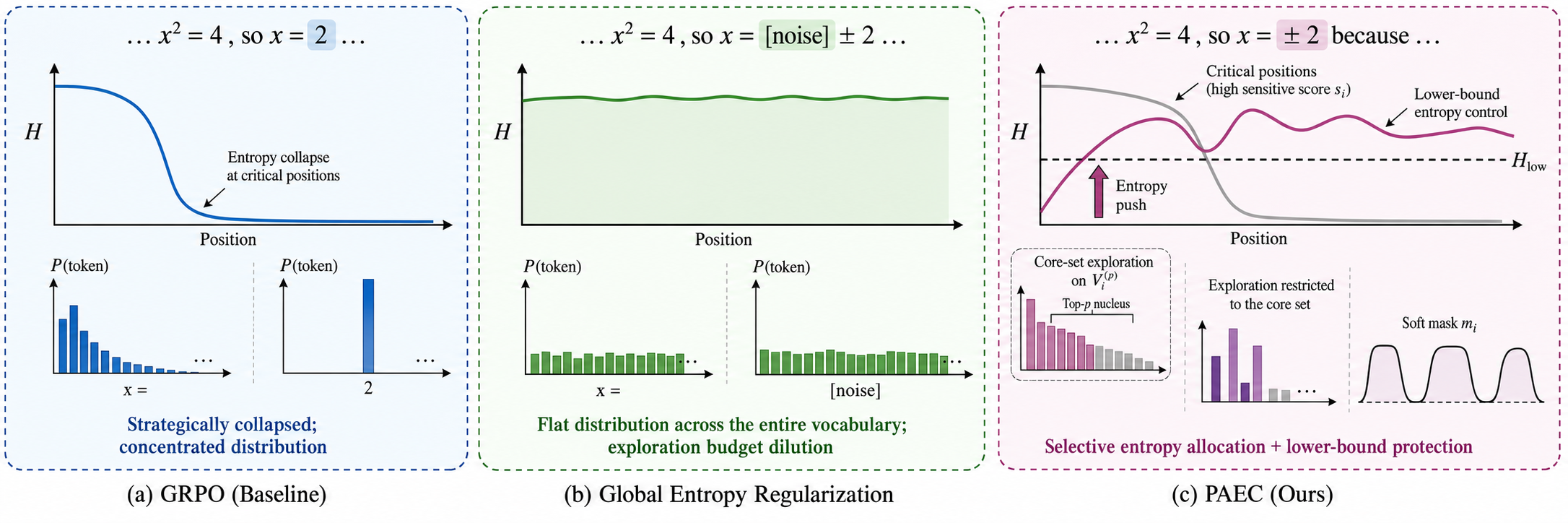} % 宽度、图片名
  \caption{Conceptual comparison of GRPO, Global Entropy Regularization, and PAEC. GRPO may suffer from entropy collapse, global entropy regularization spreads exploration uniformly, while PAEC selectively calibrates entropy at decision-sensitive positions with an anchor-based lower-bound penalty.}
\end{figure*}

\section{Introduction}
Reinforcement Learning has become a central approach for improving the complex reasoning capabilities of large language models. More recent work on verifiable tasks, such as mathematics and code, has promoted reinforcement learning with verifiable rewards (RLVR), where answer correctness can be directly used as a training signal~\citep{schulman2017proximalpolicyoptimizationalgorithms,Guo_2025}. Despite its empirical success, RLVR introduces a challenging exploration problem: during training, the policy distribution often sharpens rapidly, and the model may prematurely concentrate on a small set of high-frequency reasoning paths. This phenomenon is commonly referred to as entropy collapse~\citep{jiang2025rethinkingentropyregularizationlarge,chen2026explorationvsexploitationrethinking}. If the policy becomes overconfident at some critical positions which can strongly influence the subsequent trajectory, the model may lose access to alternative solution branches and instead reinforce a narrow set of habitual patterns.

Recent analyses further suggest that RLVR may not always expand the reasoning frontier of the base model. Fine-grained Pass@K analyses indicate that, at large K, RL-trained models can perform comparably to or even worse than their base models~\citep{yue2025doesreinforcementlearningreally}. This suggests that RLVR may, in some settings, primarily reallocate probability mass within the reasoning support already present in the base model, rather than reliably creating fundamentally new reasoning capabilities. From this perspective, RL can make the model more likely to sample existing high-confidence paths, thereby sharpening the policy distribution and accelerating entropy collapse~\citep{karan2025reasoningsamplingbasemodel}. 

A natural way to mitigate entropy collapse is to add entropy regularization. Maximum entropy reinforcement learning shows that entropy can provide an additional exploration incentive beyond reward maximization, and adaptive temperature updates can improve training stability~\citep{haarnoja2019softactorcriticalgorithmsapplications}.
However, directly applying global entropy regularization to long-form reasoning is inefficient. Language models operate over extremely large discrete vocabularies, and uniformly increasing entropy can allocate probability mass to low-confidence tail tokens. Moreover, reasoning trajectories contain many formatting, connective, and rhetorical tokens that have limited influence on downstream solution branches. 
Thus, the central question in reasoning-oriented RLVR is not only how much entropy should be encouraged, but also which token positions should receive the limited exploration budget.

Recent work has begun to improve entropy management in RLVR from different perspectives. Some methods modify policy-gradient updates or clipping behavior to preserve exploration-related gradients, while others introduce explicit entropy terms. Adaptive Entropy Regularization (AER), shows that entropy coefficients should not remain fixed, and instead adjusts the strength of regularization according to task difficulty and training state~\citep{zhang2026revisitingentropyregularizationadaptive}. These methods provide important insights, but they mainly decide how much entropy regularization to apply at the sample or sequence level. They do not explicitly address which token positions within a long reasoning trajectory should receive the exploration budget.

Based on this observation, we propose \textbf{Position-Aware Entropy Calibration} (PAEC), a token-position-level entropy calibration method for RLVR. PAEC avoids uniformly increasing entropy over the entire response. Instead, it first constructs a position-aware score for each generated token by combining normalized local entropy over a truncated top-$p$ distribution with the local top-two log-probability gap. This score is used to derive a soft mask, which allocates selective entropy regularization to positions that are more likely to affect subsequent reasoning branches. In addition, PAEC introduces an initial-anchor-based lower-bound entropy penalty. When the aggregated entropy over high-score positions falls below a reference threshold, the penalty is activated to compensate for low-entropy states; when entropy remains sufficiently high, the penalty stays inactive. Importantly, PAEC does not attempt to identify "true" critical tokens in a causal sense. Instead, it constructs a lightweight decision-sensitive proxy that better matches the structure of reasoning tasks. Our main contributions are as follows:
\begin{itemize}
    \item We propose a position-aware selective entropy mechanism that restricts entropy computation to a top-$p$ candidate set and constructs a soft mask from normalized local entropy and local top-two log-probability competition. This enables entropy regularization to focus on positions more likely to influence downstream reasoning branches.
    \item We design an anchor-based lower-bound entropy penalty that activates only when the masked entropy falls below an initial reference level, providing targeted compensation for late-stage entropy collapse.
    \item Experiments on five mathematical reasoning benchmarks show that PAEC improves macro-average majority-vote performance over GRPO, Global Entropy Regularization, CISPO, Clip-Cov/KL-Cov, and AER.
\end{itemize}

\section{Related Work}
\subsection{RLVR}
Early reinforcement learning from human feedback (RLHF) emphasizes aligning model outputs with human preferences, with representative preference-optimization methods including DPO~\citep{rafailov2024directpreferenceoptimizationlanguage} and KTO~\citep{ethayarajh2024ktomodelalignmentprospect}. In the context of improving large language model reasoning, Proximal Policy Optimization (PPO)~\citep{schulman2017proximalpolicyoptimizationalgorithms} and its variants have become a core technical foundation for RLVR. PPO stabilizes policy updates by combining importance sampling with a clipping mechanism. Its objective is defined as:
\begin{equation}
\begin{aligned}
L(\theta)
= \hat{\mathbb{E}}_t \Big[
\min \big(
&r_t(\theta)\hat{A}_t, \\
&\mathrm{clip}(r_t(\theta), 1-\epsilon, 1+\epsilon)\hat{A}_t
\big)
\Big].
\end{aligned}
\end{equation}
Here, $r_t(\theta)$ denotes the importance sampling ratio. In the autoregressive setting, it can be written as $\frac{
\pi_\theta(o_{i,t}\mid q,o_{i,<t})
}{
\pi_{\theta_{\mathrm{old}}}(o_{i,t}\mid q,o_{i,<t})
}$, and $\hat{A}_t$ denotes the GAE advantage estimate. Although PPO is a mature policy-gradient method, it typically relies on a value model for advantage estimation, which increases training complexity. To improve practical efficiency, DeepSeek introduced Group Relative Policy Optimization (GRPO)~\citep{shao2024deepseekmathpushinglimitsmathematical}, which simplifies PPO by removing the independent value network and replacing GAE with group-relative advantage estimation. Specifically, the advantage of each response is computed by subtracting the group mean reward and normalizing by the group reward standard deviation, substantially reducing training overhead. In practice, the reward model is often replaced by a binary reward signal; for example, in mathematical reasoning, a response receives reward 1 if the final answer is correct and 0 otherwise. The GRPO objective is:
\begin{equation}
\begin{aligned}
\mathcal{J}_{\mathrm{GRPO}}(\theta)
&=
\frac{1}{G}
\sum_{i=1}^{G}
\frac{1}{|o_i|}
\sum_{t=1}^{|o_i|}
\Big[
\min \big(
\rho_{i,t}(\theta)\hat{A}_{i,t},
\\
&\quad
\mathrm{clip}(\rho_{i,t}(\theta),1-\epsilon,1+\epsilon)
\hat{A}_{i,t}
\big)
\\
&\quad
-
\beta_{\mathrm{KL}}
D_{\mathrm{KL}}
\big(
\pi_\theta
\Vert
\pi_{\mathrm{ref}}
\big)
\Big].
\end{aligned}
\end{equation}
where $\beta$ is the KL penalty coefficient used to constrain the policy update. Although GRPO improves training efficiency, it still suffers from policy entropy collapse and reduced reasoning-path diversity in reasoning tasks, which limits its performance in more complex settings.

\subsection{Entropy Collapse and Entropy Regularization}
To address entropy collapse in GRPO-style training, recent studies have explored several remedies. One line of work modifies the gradient update mechanism, such as preserving important token gradients through importance-ratio clipping. Representative examples include the Clip-Higher strategy in DAPO~\citep{yu2025dapoopensourcellmreinforcement} and CISPO~\citep{minimax2025minimaxm1scalingtesttimecompute}. Other recent methods mitigate entropy collapse by suppressing updates on high-covariance tokens~\citep{cui2025entropymechanismreinforcementlearning}, or by adding a plug-and-play, gradient-detached entropy term to the advantage component of the objective~\citep{cheng2025reasoningexplorationentropyperspective}. Another line of work introduces explicit entropy regularization, i.e., global entropy regularization, which adds an entropy term to the objective and encourages policy stochasticity through entropy maximization~\citep{haarnoja2018softactorcriticoffpolicymaximum}. The token-level entropy is defined as:
\begin{equation}
\mathcal{H}_j = -\sum_{v \in \mathcal{V}} \pi_\theta(v \mid q, v_{<j}) \log \pi_\theta(v \mid q, v_{<j}).
\end{equation}
The resulting objective can be written as:
\begin{equation}
\mathcal{J}(\theta) = \mathcal{J}_{\text{PO}}(\theta) + \alpha \cdot \mathbb{E}\left[H\left(\pi_\theta(\cdot \mid s_t)\right)\right]
\end{equation}
where $\alpha$ is the entropy regularization coefficient that balances reward maximization and exploration, and $\mathcal{V}$ denotes the full output vocabulary of the language model.

Although these methods have achieved substantial improvements, they do not explicitly determine where the entropy budget should be allocated. Recent work has also identified this limitation~\citep{jiang2025rethinkingentropyregularizationlarge} and proposed selective entropy regularization. However, directly discarding "non-critical" tokens may remove useful entropy-gradient information from some positions.

\subsection{Adaptive Entropy Regularization}
Inspired by maximum-entropy reinforcement learning, recent work has also studied how to adaptively adjust the entropy regularization coefficient. A representative example is Adaptive Entropy Regularization (AER), which revisits entropy regularization in RLVR~\citep{zhang2026revisitingentropyregularizationadaptive}. AER shows that a fixed entropy coefficient may be unstable across models, datasets, and training stages, and adaptively controls the strength of entropy regularization through difficulty-aware coefficient allocation, initial-anchored target entropy, and dynamic global coefficient updates. Similar ideas have also been adopted in the reinforcement-learning post-training of Skywork Open Reasoner 1~\citep{he2025skyworkopenreasoner1}.

PAEC is complementary to this line of work: rather than only deciding how much entropy regularization should be applied at the sample or sequence level, PAEC studies where the entropy budget should be allocated within a long reasoning trajectory. 
This position-level perspective is particularly important for mathematical reasoning, where only a small subset of local decisions may determine subsequent solution branches.

\section{Methodology}
In this section, we present the full PAEC framework and describe its training objective in detail.
\subsection{Position-Aware Soft Mask}
The first component of PAEC is a position-aware soft mask. We first construct a top-$p$ policy nucleus at each token position, and then generate a soft mask $m_i$ from a position-aware score $s_i$ along the trajectory.

At generation position $i$, let 
$\pi_i(v) \equiv \pi_\theta(v \mid q, o_{<i})$
denote the next-token distribution over the vocabulary $\mathcal{V}$.
Let $v^{(1)}, v^{(2)}, \ldots$ be the vocabulary tokens sorted in descending order by $\pi_i(v)$.
We define the top-$p$ policy nucleus as the smallest prefix whose cumulative probability mass reaches $p$:
$k_i = \min \left\{ k: \sum_{\ell=1}^{k} \pi_i(v^{(\ell)}) \ge p \right\}, 
\mathcal{V}^{(p)}_i = \{v^{(1)}, \ldots, v^{(k_i)}\}.$

The corresponding probability mask is 
\[
M_i^{p}(v)=
\begin{cases}
1, & v\in \mathcal{V}_i^{(p)},\\
0, & \text{otherwise}.
\end{cases}
\]
Based on this mask, we define the renormalized top-$p$ distribution:
\begin{equation}
\pi'_i(v) = \frac{M_i^{p}(v) \cdot \pi_i(v)}{\sum_{u\in\mathcal{V}_i^{(p)}}M_i^{p}(u) \cdot \pi_i(u)},
\end{equation}
and compute the local entropy:
\begin{equation}
H'_i = -\sum_{v\in\mathcal{V}_i^{(p)}} \pi'_i(v)\log \pi'_i(v).
\end{equation}
Here, both $\pi'_i(v)$ and $H'_i$ are defined only on the policy nucleus, so entropy is restricted to the candidate subset that is currently regarded as semantically plausible by the policy.

Because the size of the top-$p$ candidate set varies across positions, directly using $H'_i$ would confound uncertainty with candidate-set size. Noting that the maximum entropy of a discrete distribution over $|\mathcal{V}_i^{(p)}|$ elements is $\log |\mathcal{V}_i^{(p)}|$, we first normalize the local entropy as
\begin{equation}
s_1 = \frac{H'_i}{\log |\mathcal{V}_i^{(p)}| + \epsilon}.
\end{equation}
The normalized score $s_1\in[0,1]$ measures the relative uncertainty of the current position within its top-$p$ policy nucleus.

Next, let $\pi_i^{(1)}$ and $\pi_i^{(2)}$ denote the top two probabilities under the renormalized distribution $\pi'_i(v)$. We define the log-gap between the top two candidates as $\Delta_i = \log \pi_i^{(1)} - \log \pi_i^{(2)}$. Based on this quantity, we define the decision-competition score:
\begin{equation}
s_2 = 2 \cdot \sigma(-\Delta_i).
\end{equation}
$\sigma(·)$ is the sigmoid function. The factor 2 rescales the maximum value to 1 when the top two candidates are tied. When $\Delta_i$ is small, the top two candidates are closer to each other, indicating a more ambiguous decision boundary; consequently, $s_2$ becomes larger. We combine the normalized entropy score and the decision-competition score into a detached position-aware score:
\begin{equation}
s_i = \text{sg}\bigl(s_1 + \beta s_2\bigr).
\end{equation}
where $\beta\in[0,1]$ controls the contribution of local candidate competition and $\mathrm{sg}(\cdot)$ denotes stop-gradient. The stop-gradient operation ensures that the position-aware soft mask is used only to select where entropy regularization is applied, rather than providing an additional optimization path through which the model can manipulate the mask itself.

Let τ denote the $\rho$-quantile of the score sequence $\{s_i\}_{i=1}^{|o|}$. That is, $\tau=\mathrm{Quantile}_{\rho}(\{s_i\}_{i=1}^{|o|}).$

The position-aware soft mask is then defined as
\begin{equation}
m_i = \sigma\bigl(\kappa(s_i - \tau)\bigr).
\end{equation}
where $\kappa$ > 0 is a scaling factor. Since $s_i$ is already detached by sg(·), and τ is computed from the detached score sequence, $m_i$ does not backpropagate gradients to the policy parameters. In this way, the soft mask acts purely as a position selector rather than an additional optimization channel. A gradient analysis is provided in Appendix~A.

For a batch containing B responses, the aggregated entropy over critical tokens is defined as
\begin{equation}
\bar{H}(\theta) = \frac{\sum_{b=1}^{B}\sum_{i=1}^{|o_i|} m_{b,i}H'_{b,i}}{\sum_{b=1}^{B}\sum_{i=1}^{|o_i|} m_{b,i}+\epsilon},
\end{equation}
where $\epsilon$ is a numerical stabilizer. The resulting $\bar{H}(\theta)$ measures the average exploration level over token positions assigned larger mask weights.

\subsection{Lower-Bound Entropy Penalty}
After determining where entropy should be applied, we further regulate the strength of exploration during training. To this end, we adopt an entropy-budget control mechanism based on initial entropy anchoring and a lower-bound entropy penalty. Intuitively, when the average entropy over the selected positions remains sufficiently high, no additional correction is required; when it falls below a predefined lower bound, the penalty activates to prevent premature entropy collapse.

We first define an initial entropy anchor $H_0$. During the first K update steps, we initialize it using the average aggregated entropy:
\begin{equation}
H_0 = \frac{1}{K} \sum_{t=1}^K \bar{H}_t(\theta).
\end{equation}
This initialization reflects the natural uncertainty of the model before accuracy-driven reinforce ment learning becomes dominant, and provides a model and data dependent reference point for subsequent entropy control.

Based on this anchor, we define the entropy lower bound as $H_{\text{low}} = \rho_{min} H_0, \quad \rho_{min} \in (0, 1).$ Here, $\rho_{min}$ controls the minimum acceptable entropy level. As long as $\bar{H}(\theta)$ stays above Hlow, no additional entropy penalty is imposed. Once $\bar{H}(\theta)$ falls below this threshold, the penalty becomes active.

Specifically, we define the lower-bound entropy penalty as
\begin{equation}
L_{\text{penalty}} = \left[ \max\left(0, H_{\text{low}} - \bar{H}(\theta)\right) \right]^2.
\end{equation}
This one-sided penalty has a clear stage-wise interpretation: during the early training phase, the policy entropy is typically high, so the penalty remains inactive; as training proceeds, if the average entropy over critical tokens drops below the lower bound, $L_{\text{penalty}}$ increases and explicitly penalizes over-confident low-entropy states.

\subsection{Overall Objective and Training}
Combining the components above, PAEC applies two complementary entropy forces during training: on the one hand, the position-aware soft mask maintains necessary exploration on critical positions; on the other hand, the entropy lower-bound penalty provides a corrective signal whenever the entropy becomes too small. The final objective is therefore written as
\begin{equation}
J_{\text{PAEC}}(\theta) = J_{\text{GRPO}}(\theta) + \alpha \bar{H}(\theta) - \lambda_{\text{ent}} L_{\text{penalty}}.
\end{equation}
where $\lambda_{\text{ent}}$ > 0 is the coefficient of the lower-bound entropy penalty. Equivalently, under a loss-minimization implementation, we can write
\begin{equation}
L_{\text{PAEC}}(\theta) = L_{\text{GRPO}}(\theta) - \alpha \bar{H}(\theta) + \lambda_{\text{ent}} L_{\text{penalty}}.
\end{equation}
PAEC reallocates the exploration budget from uniform distribution over the entire vocabulary and trajectory to targeted allocation over critical decision tokens. The complete training procedure is summarized in Algorithm 1 in Appendix~A.
\begin{table*}[t]
\centering
\setlength{\tabcolsep}{4pt}
\renewcommand{\arraystretch}{1.2}
\resizebox{\textwidth}{!}{%
\begin{tabular}{l|cc|cc|cc|cc|cc|cc}
\hline
\multirow{2}{*}{Method} & \multicolumn{2}{c|}{AIME24} & \multicolumn{2}{c|}{AIME25} & \multicolumn{2}{c|}{AIME26} & \multicolumn{2}{c|}{MATH500} & \multicolumn{2}{c|}{AMC23} & \multicolumn{2}{c}{Average} \\
 & Maj@32 & Avg@32 & Maj@32 & Avg@32 & Maj@32 & Avg@32 & Maj@8 & Avg@8 & Maj@16 & Avg@16 & Maj & Avg \\
\hline
Qwen2.5-Math-1.5B & 16.6 & 9.2 & 13.3 & 4.8 & 10.0 & 6.3 & 45.8 & 32.7 & 57.5 & 36.2 & 28.6 & 17.8 \\
\hline
GRPO & 20.0 & 13.1 & 10.0 & 6.3 & 10.0 & 6.3 & 72.6 & 68.5 & \textbf{65.0} & 57.3 & 35.5 & 30.3 \\
CISPO & 23.3 & 16.2 & 16.6 & 9.3 & 13.3 & 8.9 & 75.0 & 70.7 & 60.0 & 56.2 & 37.6 & \textbf{32.3} \\
Global Entropy Reg & \textbf{26.7} & \textbf{16.3} & 13.3 & 7.9 & 9.8 & 9.0 & 75.4 & 68.7 & 60.0 & 54.0 & 37.0 & 31.1 \\
Clip-cov & 23.3 & 15.9 & 20.0 & 10.7 & 9.5 & 6.5 & 73.6 & 69.2 & 62.5 & 55.7 & 37.8 & 31.6 \\
KL-cov & 23.3 & 14.0 & 20.0 & 9.8 & 10.0 & 7.1 & 73.2 & 69.7 & 62.5 & \textbf{57.5} & 37.8 & 31.6 \\
AER & 20.0 & 14.9 & 23.3 & \textbf{14.3} & 16.7 & 9.1 & 70.8 & 66.6 & 62.5 & \textbf{58.6} & 38.7 & 32.4 \\
PAEC(Ours) & \textbf{26.7} & 13.6 & \textbf{26.7} & 11.8 & \textbf{16.7} & \textbf{9.3} & \textbf{75.6} & \textbf{71.0} & 62.5 & 58.0 & \textbf{41.6} & \textbf{32.7} \\
\hline
\end{tabular}
}
\raggedright\caption{Main results on five mathematical reasoning benchmarks. Due to computational constraints, benchmark-size and difficulty differences, we use K=32 for AIME-style benchmarks, K=16 for AMC23, and K=8 for MATH500.}
\label{tab:main_results}
\end{table*}

\section{Experiments}
\subsection{Experimental Setup}
\paragraph{Training Details.}We use Qwen2.5-Math-1.5B as the primary backbone in the main comparison. All methods are trained for 720 update steps with a total batch size of 96, a PPO minibatch size of 24, and 8 sampled responses per prompt. The maximum prompt and response lengths are set to 512 and 3072 tokens, respectively. We save checkpoints every 10 steps and select the checkpoint with the best validation Pass@16 on AIME23 for final evaluation. We train all methods on DAPO-Math-17K~\citep{yu2025dapoopensourcellmreinforcement}. All methods are implemented in the VeRL reinforcement learning framework~\citep{sheng2024hybridflow} under the same training budget. Specific training details are provided in Appendix~B.
\paragraph{Baselines.}We compare PAEC against several representative baselines. These include vanilla GRPO, Global Entropy Regularization, CISPO~\citep{minimax2025minimaxm1scalingtesttimecompute}, and the covariance-based entropy-control methods Clip-cov and KL-cov~\citep{cui2025entropymechanismreinforcementlearning}. We additionally include Adaptive Entropy Regularization (AER), a recent adaptive entropy baseline that dynamically adjusts entropy coefficients using difficulty-aware allocation, initial-anchored target entropy, and global coefficient updates. To ensure a fair comparison, we implement AER under the same GRPO backbone, training budget, rollout number, maximum sequence length, and evaluation protocol as PAEC. We report the AER hyperparameters in Appendix~B.
\paragraph{Benchmarks.}
We evaluate all methods on five public mathematical reasoning benchmarks:
AIME24, AIME25, AIME26~\citep{aime24,aime25,aime26},
MATH500~\citep{hendrycks2021measuringmathematicalproblemsolving}, and AMC23. These benchmarks cover different difficulty levels and are commonly used to evaluate mathematical reasoning in large language models.

\subsection{Main Results}
Table~\ref{tab:main_results} presents the main comparison across five mathematical reasoning benchmarks. PAEC achieves the best macro-average majority-vote performance among all compared methods, with an Average Maj score of 41.6. This improves over GRPO by 6.1 points, CISPO by 4.0 points, Clip-cov/KL-cov by 3.8 points, and AER by 2.9 points. In terms of Average Avg, PAEC obtains 32.7, which is slightly higher than AER (32.4) and CISPO (32.3), but the margin is relatively small. Therefore, the main advantage of PAEC lies in improving majority-vote consistency rather than uniformly dominating per-sample average accuracy.

The results indicate that PAEC does not dominate every individual metric, but it yields the strongest overall average performance and shows clearer gains on the more challenging benchmarks. This behavior is consistent with the design motivation of PAEC: rather than uniformly increasing randomness across the entire response, it reallocates exploration toward positions that are more likely to influence the downstream reasoning trajectory.

\section{Analysis and Discussion}
\begin{table*}[t]
\centering
\setlength{\tabcolsep}{4pt}
\renewcommand{\arraystretch}{1.2}
\resizebox{\textwidth}{!}{%
\begin{tabular}{l|cc|cc|cc|cc|cc|cc}
\hline
\multirow{2}{*}{Setting} & \multicolumn{2}{c|}{AIME24} & \multicolumn{2}{c|}{AIME25} & \multicolumn{2}{c|}{AIME26} & \multicolumn{2}{c|}{MATH500} & \multicolumn{2}{c|}{AMC23} & \multicolumn{2}{c}{Average} \\
 & Maj@32 & Avg@32 & Maj@32 & Avg@32 & Maj@32 & Avg@32 & Maj@8 & Avg@8 & Maj@16 & Avg@16 & Maj & Avg \\
\hline
PAEC & 26.7 & 13.6 & 26.7 & 11.8 & 16.7 & 9.3 & 75.6 & 71.0 & 62.5 & 58.0 & \textbf{41.6} & \textbf{32.7} \\
\hline
GRPO & 20.0 & 13.1 & 10.0 & 6.3 & 10.0 & 6.3 & 72.6 & 68.5 & 65.0 & 57.3 & 35.5 & 30.3 \\
w/o Lower-Bound Entropy Penalty & 23.3 & 16.6 & 20.0 & 8.7 & 16.7 & 9.3 & 71.6 & 66.8 & 62.5 & 55.0 & 38.8 & 31.2 \\
w/o Soft Mask & 20.0 & 13.4 & 23.3 & 11.4 & 13.3 & 8.2 & 75.2 & 69.7 & 60.0 & 57.6 & 38.3 & 32.0 \\
\hline
\end{tabular}%
}
\caption{The overall performance of the ablation experiment on the five benchmarks.}
\label{tab:ablation}
\end{table*}

\subsection{Ablation Study}
Table~\ref{tab:ablation} evaluates the contribution of the two PAEC components. Removing either component degrades the full model, confirming that PAEC benefits from both position-aware allocation and entropy-floor correction. Without the lower-bound entropy penalty, the macro-average Maj score drops from 41.6 to 38.8, suggesting that the soft mask alone is insufficient to prevent selected-position entropy from continuing to shrink during later training. Without the soft mask, the macro-average Maj further drops to 38.3, indicating that applying the entropy floor without position-aware allocation weakens the majority-vote consistency. Interestingly, the w/o Soft Mask variant obtains a slightly higher Avg score than w/o Lower-Bound Entropy Penalty, which suggests that the two components may affect different aspects of generation: the entropy-floor penalty helps maintain per-sample exploration, whereas the soft mask improves how this exploration is allocated across reasoning positions.
% 熵曲线
% Main training dynamics
\begin{figure*}[t]
  \centering
  \captionsetup[subfigure]{labelformat=simple,labelsep=space}
  \renewcommand\thesubfigure{(\alph{subfigure})}
  \begin{subfigure}[t]{0.48\textwidth}
    \centering
    \includegraphics[width=\linewidth]{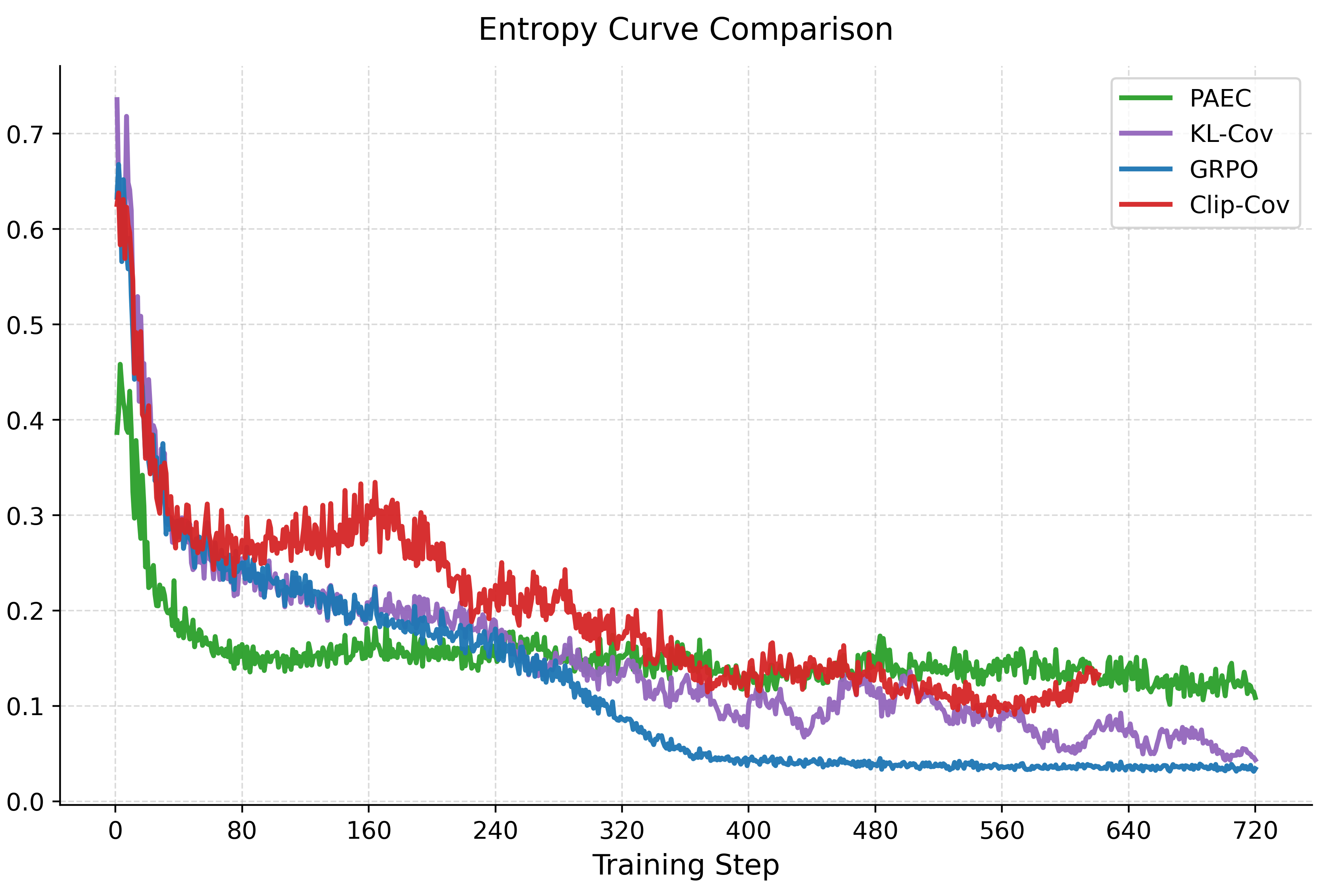}
    \caption{Training-time policy entropy of different RLVR methods.}
    \label{fig:entropy_curve}
  \end{subfigure}
  \hfill
  \begin{subfigure}[t]{0.48\textwidth}
    \centering
    \includegraphics[width=\linewidth]{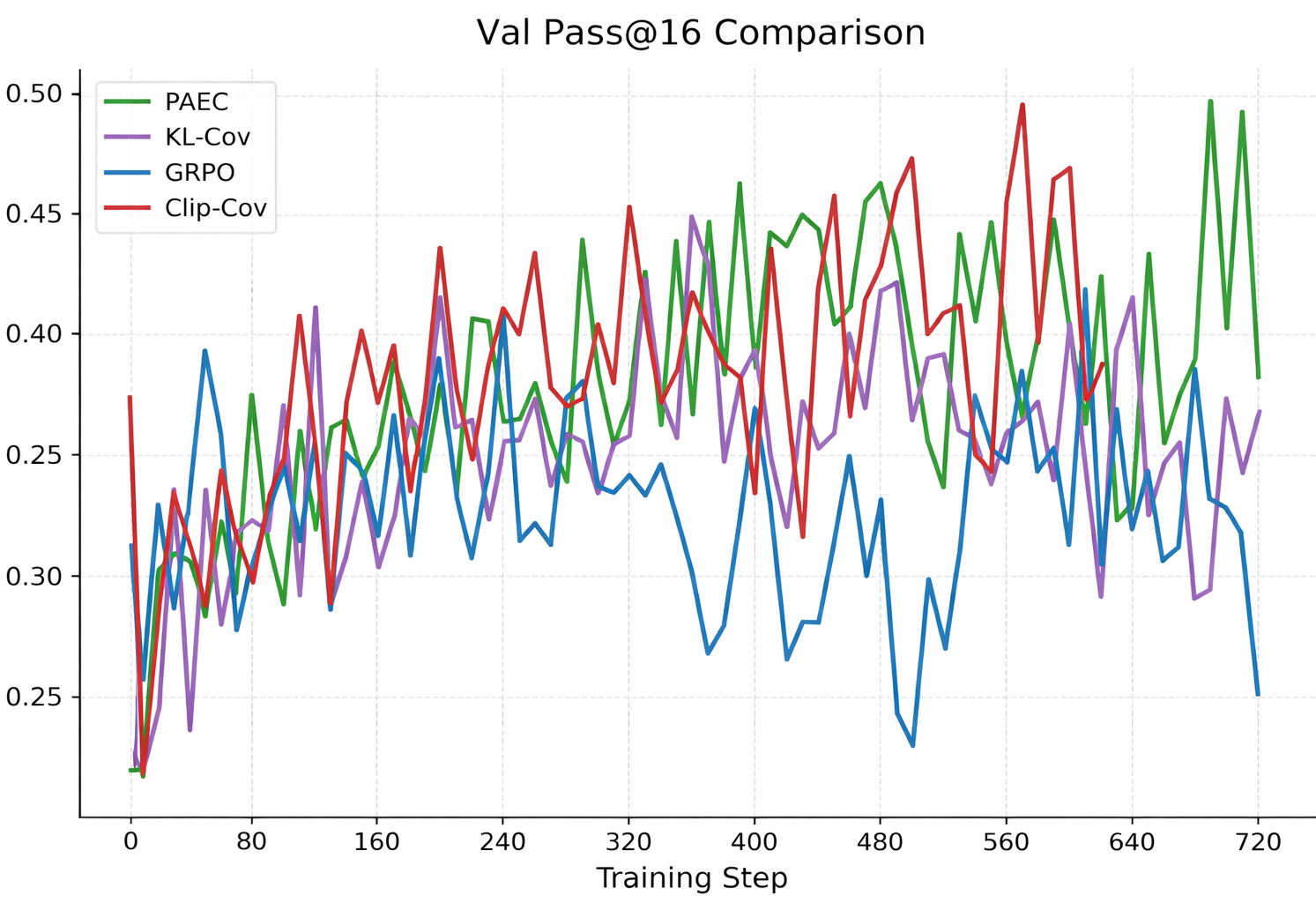}
    \caption{Validation Pass@16 curves of different RLVR methods.}
    \label{fig:pass16_curve}
  \end{subfigure}
  \caption{Main training dynamics during RLVR training. PAEC maintains a moderate entropy level while preserving competitive validation performance. Additional mask and response-length diagnostics are reported in Appendix~\ref{app:experimental_details}.}
  \label{fig:training_dynamics}
\end{figure*}

\subsection{Entropy Dynamics during Training}
Figure~\ref{fig:entropy_curve} plots the training-time policy entropy of GRPO, Clip-cov, KL-cov, and PAEC. All methods show a rapid entropy decrease at the beginning of RLVR training, indicating that reward optimization quickly sharpens the policy distribution. However, the later-stage behavior differs substantially. The entropy of GRPO continues to decline and reaches a very low level, suggesting stronger policy concentration. In contrast, PAEC stabilizes the entropy around a moderate level in the later stage. This pattern is consistent with the design of the entropy-floor penalty: the penalty is inactive when the selected-position entropy is sufficiently high, but becomes active once the aggregated masked entropy falls below the anchor-based lower bound.

Figure~\ref{fig:pass16_curve} further shows that this entropy stabilization does not come at the cost of validation performance. PAEC achieves a validation Pass@16 curve that is competitive with Clip-cov while being less volatile than vanilla GRPO. Therefore, PAEC does not simply increase randomness; instead, it maintains a moderate level of exploration while preserving reward-driven learning. Additional training diagnostics, including the mask-mean curve and response-length curve, are reported in Appendix~\ref{app:experimental_details}.

\subsection{Where Does PAEC Allocate Entropy?}
To inspect the allocation behavior of the soft mask, we visualize mask values along generated reasoning trajectories. In the case study in Appendix~C, PAEC assigns relatively high mask weights to tokens such as \texttt{Therefore}, \texttt{Since}, \texttt{Let}, and \texttt{implies}, which often appear near local reasoning transitions, algebraic transformations, or conclusion-forming steps. This observation is consistent with recent evidence that high-entropy minority tokens can play an important role in effective RLVR updates~\citep{wang20258020rulehighentropyminority}.

We further report the aggregate mask-mean curve in Appendix~\ref{app:experimental_details}. The mean mask value gradually increases from roughly 0.32 to around 0.45, indicating that PAEC maintains non-uniform but non-degenerate position weights. Since the mask is continuous, this value should be interpreted as average weighting strength rather than the fraction of selected tokens. This suggests that PAEC neither collapses into a hard sparse selector nor becomes uniform global entropy regularization.

\subsection{Sampling-Based Performance and Response Length}
Beyond majority-vote accuracy, we further examine Pass@K to evaluate sampling-dependent solution coverage. Unlike Maj@K, which emphasizes majority-vote consistency, Pass@K measures whether the model can sample at least one correct solution under a fixed decoding distribution. The complete Pass@K curves on AIME25 and MATH500 are reported in Appendix~\ref{app:experimental_details}. PAEC is stronger on both, indicating that its gains extend beyond majority voting.

We also analyze response length as an auxiliary behavioral diagnostic, with the full curve reported in Appendix~\ref{app:experimental_details}. All methods initially produce shorter responses under the 3072-token rollout budget, while entropy-preserving methods, including PAEC, exhibit a mild later-stage length recovery. We therefore treat response length as a diagnostic of generation behavior rather than direct evidence of reasoning quality.

\begin{figure}[t]
  \centering
  \includegraphics[width=\columnwidth]{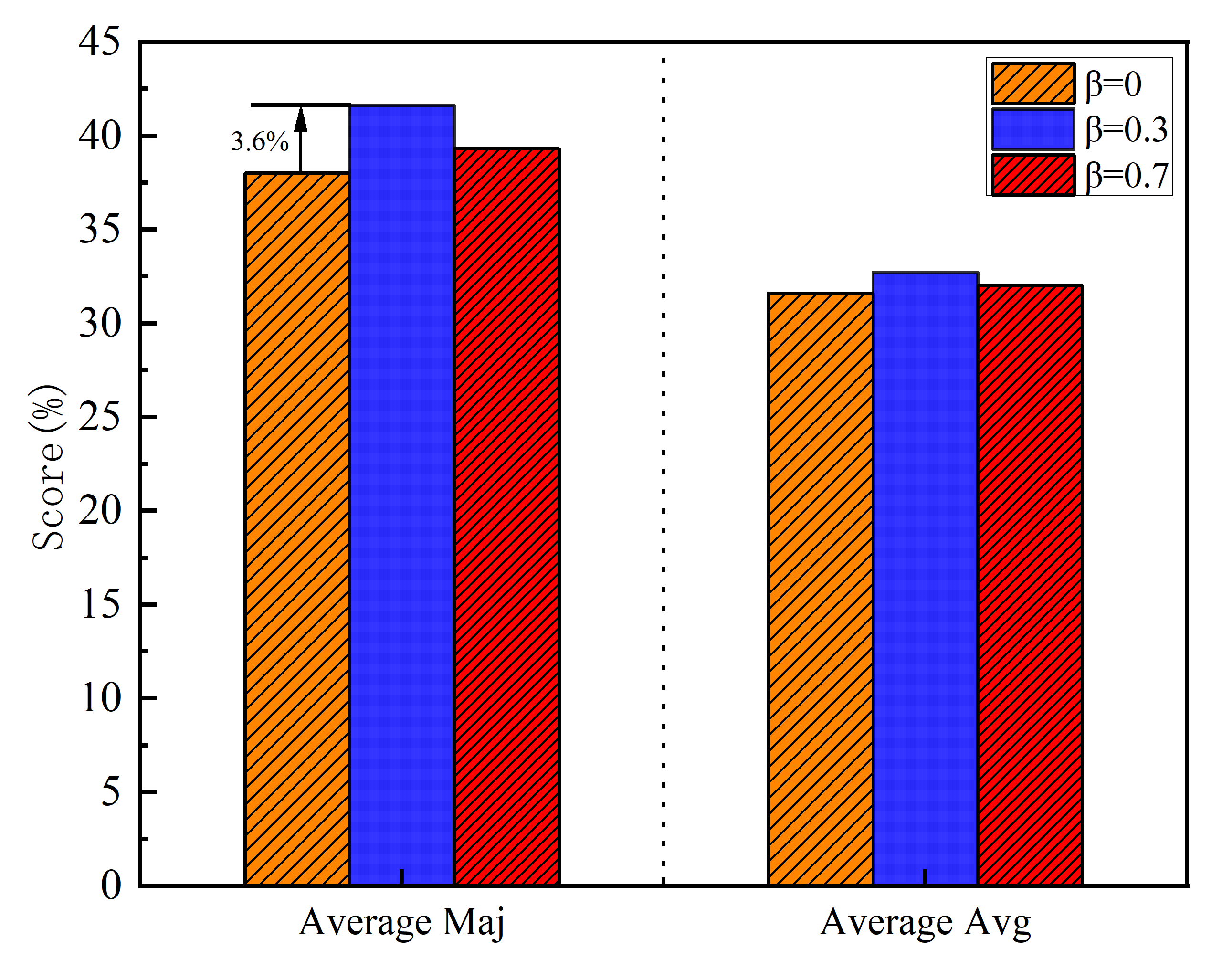}
  \caption{Effect of the competition weight $\beta$ on PAEC. 
When $\beta=0$, the position-aware score only uses the normalized local entropy score $s_1$. 
A moderate value $\beta=0.3$ achieves the best Average Maj and Average Avg, while a larger value $\beta=0.7$ slightly degrades performance.}
  \label{fig:beta}
\end{figure}

\begin{figure}[t]
  \centering
  \includegraphics[width=\columnwidth]{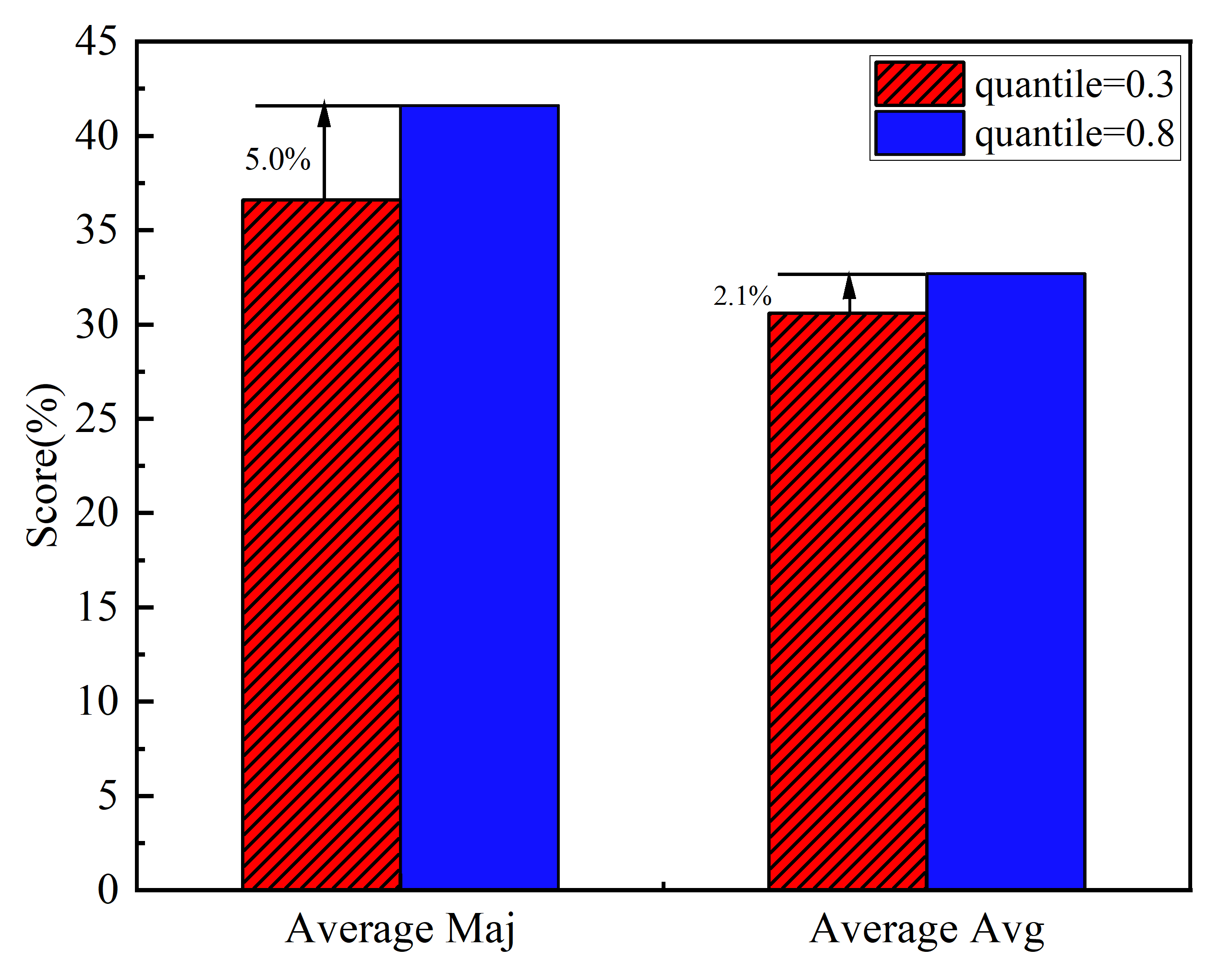}
  \caption{Effect of the quantile threshold $\rho$ on mask selectivity; a higher quantile focuses the entropy budget on fewer high-score positions.}
  \label{fig:quantile}
\end{figure}

\subsection{Hyperparameter Settings and Robustness Analysis}
Finally, we analyze the sensitivity of PAEC to two mask-related hyperparameters: the competition weight $\beta$ and the quantile threshold $\rho$ used for constructing the position-aware soft mask. Figure~\ref{fig:beta} compares different values of $\beta$, which controls the contribution of the top-two competition score $s_2$ in the position-aware score. When $\beta=0$, the mask score only depends on the normalized local entropy score $s_1$, without incorporating the local competition between the top two candidate tokens. 
This setting obtains an Average Maj of 38.0 and an Average Avg of 31.6. Adding a moderate competition term with $\beta=0.3$ improves the Average Maj to 41.6 and the Average Avg to 32.7, yielding gains of 3.6 and 1.1 points, respectively. 
This indicates that top-two candidate competition provides useful complementary information beyond local entropy alone. However, further increasing $\beta$ to 0.7 reduces the Average Maj to 39.3 and the Average Avg to 32.0. 
This suggests that over-emphasizing the competition score may weaken the dominant role of normalized local entropy. Therefore, we use $\beta=0.3$ as the default setting, which provides a balanced combination of local entropy and candidate competition.

The Figure~\ref{fig:quantile} compares different quantile thresholds for the soft mask. Using a higher quantile threshold, $\rho=0.8$, substantially outperforms $\rho=0.3$. The average Maj improves from 36.6 to 41.6, and the average Avg improves from 30.6 to 32.7. This supports the selective-allocation motivation of PAEC: when the threshold is too low, too many positions receive non-negligible entropy weights, making the method closer to global entropy regularization and diluting the exploration budget. A higher quantile preserves stronger selectivity and focuses the entropy bonus on high-score positions. The specific settings of other parameters and the related precautions will be provided in detail in Appendix~B.

Overall, the robustness analysis shows that PAEC's strongest performance is obtained with a moderately weighted competition term and a relatively selective mask threshold, supporting our view that entropy management in reasoning RL should be selective rather than uniform.

\section{Conclusion}
This paper revisits entropy regularization in RLVR for mathematical reasoning and argues that effective exploration should be treated as a structured allocation problem over token positions rather than a uniform increase in randomness across the entire response. We introduce Position-Aware Entropy Calibration (PAEC), which uses a position-aware soft mask and an anchor-based lower-bound entropy penalty to selectively maintain exploration at decision-sensitive positions. Experiments on multiple mathematical reasoning benchmarks show that PAEC improves performance over strong RLVR baselines, highlighting the importance of selective entropy calibration for reasoning-oriented reinforcement learning.

\section*{Limitations}
This work has several limitations. First, due to computational constraints, our main experiments are conducted on a 1.5B scale model, with additional experiments on 4B- and 8B-scale models. Although the results show consistent trends in our current setting, further experiments on larger models are needed to assess whether PAEC scales to stronger reasoning models and more resource-intensive RLVR training regimes.

Second, our evaluation focuses on mathematical reasoning benchmarks. While this setting is well suited for studying RLVR because rewards can be verified automatically, it does not cover other important reasoning domains, such as code generation, theorem proving, tool-use reasoning, or multi-turn interactive reasoning. Whether position-aware entropy calibration provides similar benefits in these settings remains an open question.

Third, PAEC relies on normalized local entropy and local logit competition as a proxy for decision-sensitive positions. This proxy is simple and efficient, but it does not directly measure the causal influence of a token position on the downstream reasoning trajectory. Future work could incorporate more explicit attribution or intervention-based analyses to better understand which positions truly govern reasoning branches.

Finally, PAEC focuses on reallocating the entropy budget rather than directly controlling the policy-gradient budget. A natural extension is to use similar position-aware signals to modulate policy-gradient updates, assigning larger updates to decision-sensitive positions and smaller updates to less informative tokens. Exploring this direction may provide a more direct mechanism for controlling optimization during reasoning-oriented RLVR \citep{ma2026fipoelicitingdeepreasoning}.

\section*{Ethical Considerations}
All datasets used in this work are publicly available and contain no personal or sensitive information. Our experiments only involve open-source language models and do not collect any human subjects data. Throughout this study, we have strictly followed established ethical guidelines, ensuring that our findings are reported honestly, transparently, and with full accuracy. The proposed method, PAEC, aims to improve the exploration-exploitation trade-off during reinforcement learning for mathematical reasoning, a task that is inherently constructive and low-risk. We do not foresee direct negative societal impacts introduced by this work. We plan to release our code and training configurations under open-source licenses to facilitate reproducibility.

\nocite{*} 
\bibliography{reference}

\clearpage
\appendix
\section{Algorithms and Propositional Proofs}
\subsection{PAEC Training Algorithm}
\begin{breakablealgorithm}
\caption{PAEC Training based on GRPO}
\label{alg:paec}
\begin{algorithmic}[1]
\Require Initial policy $\pi_{\theta}$, batch size $B$, top-p threshold $p$, score weight $\beta$, quantile level $\rho$, mask sharpness $\kappa$, entropy coefficient $\alpha$, lower-bound penalty coefficient $\lambda_{\mathrm{ent}}$, anchor ratio $\rho_{min}$, anchor initialization steps $K$
\State Initialize entropy anchor accumulator $\mathcal{S}_H \gets 0$
\For{training step $t = 1,2,\dots,T$}
    \State Sample a group of responses $\{o_i\}_{i=1}^{G}$ from the old policy $\pi_{\theta_{\mathrm{old}}}$
    \State Compute verifier rewards $\{r_i\}_{i=1}^{G}$ and group-relative advantages $\{A_i\}_{i=1}^{G}$
    \For{each response $o_i$ and each token position $j$}
        \State Construct the top-$p$ nucleus $\mathcal{V}^{(p)}_{i,j}$
        \State Define the renormalized top-$p$ distribution
        \[
        \pi'_{i,j}(v)=
        \frac{M^p_{i,j}(v)\,\pi_{i,j}(v)}
        {\sum_{u\in\mathcal{V}} M^p_{i,j}(u)\,\pi_{i,j}(u)}
        \]
        \State Compute the local entropy
        \[
        H'_{i,j}=-\sum_{v\in\mathcal{V}} \pi'_{i,j}(v)\log \pi'_{i,j}(v)
        \]
        \State Compute the normalized entropy score
        \[
        s_{1,i,j}=\frac{H'_{i,j}}{\log |\mathcal{V}^{(p)}_{i,j}|+\epsilon}
        \]
        \State Let $\pi^{(1)}_{i,j}$ and $\pi^{(2)}_{i,j}$ be the top two probabilities under $\pi'_{i,j}$, and compute
        \[
        \Delta_{i,j}=\log \pi^{(1)}_{i,j}-\log \pi^{(2)}_{i,j}, 
        s_{2,i,j}=2\,\sigma(-\Delta_{i,j})
        \]
        \State Compute the detached position-aware score
        \[
        s_{i,j}=\mathrm{sg}\!\left(s_{1,i,j}+\beta s_{2,i,j}\right)
        \]
    \EndFor
    \For{each response $o_i$}
        \State Compute the score quantile threshold $\tau_i=\mathrm{Quantile}_{\rho}(\{s_{i,j}\}_{j=1}^{|o_i|})$
        \For{each token position $j$ in $o_i$}
            \State Compute the soft mask
            \[
            m_{i,j}=\sigma\!\left(\kappa(s_{i,j}-\tau_i)\right)
            \]
        \EndFor
    \EndFor
    \State Aggregate the masked critical entropy over the batch
    \[
    \bar{H}(\theta)=
    \frac{\sum_{i=1}^{B}\sum_{j=1}^{|o_i|} m_{i,j} H'_{i,j}}
         {\sum_{i=1}^{B}\sum_{j=1}^{|o_i|} m_{i,j}+\epsilon}
    \]
    \If{$t \le K$}
        \State $\mathcal{S}_H \gets \mathcal{S}_H + \bar{H}(\theta)$
        \State $H_0 \gets \mathcal{S}_H / t$
    \Else
        \State Keep $H_0$ fixed
    \EndIf
    \State Define the entropy lower bound
    \[
    H_{\mathrm{low}}=\rho_{min} H_0
    \]
    \State Compute the one-sided lower-bound penalty
    \[
    L_{\mathrm{low}}=\big[\max(0, H_{\mathrm{low}}-\bar{H}(\theta))\big]^2
    \]
    \State Compute the PAEC loss
    \[
    L_{\mathrm{PAEC}}(\theta)=L_{\mathrm{GRPO}}(\theta)-\alpha \bar{H}(\theta)+\lambda_{\mathrm{ent}}L_{\mathrm{low}}
    \]
    \State Update policy parameters:
    \[
    \theta \leftarrow \theta - \eta \nabla_{\theta} L_{\mathrm{PAEC}}(\theta)
    \]
\EndFor
\end{algorithmic}
\normalsize
\end{breakablealgorithm}

\subsection{Why Stop-Gradient Is Used}
In PAEC, the position-aware score is detached before constructing the soft mask:
\[
s_i=\mathrm{sg}(s_1+\beta s_2).
\]
This design prevents gradients from propagating through the score and the mask. Without this stop-gradient operation, the model could increase the aggregated masked entropy by changing the score values or mask weights, rather than by increasing the local entropy at the selected positions.

Let
\[
\bar{H}(\theta)=\frac{N}{Z}, 
N=\sum_i m_i H'_i, 
Z=\sum_i m_i+\epsilon.
\]
Differentiating $\bar{H}(\theta)$ with respect to $\theta$ gives
\[
\nabla_{\theta}\bar{H}
=
\frac{1}{Z}\sum_i m_i \nabla_{\theta}H'_i
+
\frac{1}{Z}\sum_i \bigl(H'_i-H\bigr)\nabla_{\theta}m_i.
\]

The first term,
\[
\frac{1}{Z}\sum_i m_i \nabla_{\theta}H'_i,
\]
is the desired component, since it adjusts the local entropy on the selected critical positions.

By contrast, the second term,
\[
\frac{1}{Z}\sum_i \bigl(H'_i-H\bigr)\nabla_{\theta}m_i,
\]
corresponds to an additional gradient path through the position score and the soft mask. If this term remains active, the model may increase the aggregated entropy by changing the position scores and mask weights, rather than by genuinely improving exploration on the selected tokens.

We now show that the single stop-gradient in $s_i$ is sufficient to remove this extra path. Since
\[
s_i=\mathrm{sg}(s_1+\beta s_2),
\]
we have
\[
\nabla_{\theta}s_i=0.
\]
Moreover, because $\tau$ is computed from the detached score sequence $\{s_i\}$, it also satisfies
\[
\nabla_{\theta}\tau=0.
\]
Applying the chain rule to the mask yields
\[
\nabla_{\theta}m_i
=
\sigma'\!\bigl(\kappa(s_i-\tau)\bigr)\cdot \kappa \cdot \nabla_{\theta}(s_i-\tau)
=
0.
\]
Substituting this into the gradient of $H(\theta)$, we obtain
\[
\nabla_{\theta}H
=
\frac{1}{Z}\sum_i m_i \nabla_{\theta}H'_i.
\]

Therefore, the stop-gradient in $s_i$ is sufficient to block the extra backward path
\[
\theta \rightarrow p'_i \rightarrow (s_1,s_2) \rightarrow s_i \rightarrow m_i \rightarrow H,
\]
preventing the model from increasing the aggregated entropy by manipulating the position scores or mask weights. As a result, the soft mask serves only to determine \emph{where} entropy should be applied, while the local entropy $H'_i$ remains the quantity that is actually optimized.

\section{Experimental Details}
\label{app:experimental_details}
\subsection{System Prompt}
We use simple and task-oriented system prompts throughout our experiments, without relying on elaborate prompt engineering. For training, we use model-specific prompts to ensure that the model produces step-by-step reasoning and places the final answer in a verifiable format. For final benchmark evaluation, we use a unified evaluation prompt across all models and methods.
\newtcolorbox{systempromptblock}[1]{
    enhanced,
    colback=green!10,         % 主体背景：极浅灰色
    colframe=green!70,       % 边框：浅灰色
    coltitle=white,
    colbacktitle=green!60!black,
    fonttitle=\bfseries,
    title={#1},
    arc=2pt,
    boxrule=0.5pt,
    breakable,
}
\begin{systempromptblock}{System prompt for Qwen model}
You are a meticulous math reasoning assistant.
Please solve the problem with clear step-by-step reasoning.
You must put the final answer in \textbackslash boxed\{...\} on the last line.
Important: Do not put anything after the boxed answer.
\end{systempromptblock}

\begin{systempromptblock}{System prompt for Llama model}
You are a meticulous math reasoning assistant. 
Please solve the problem with clear step-by-step reasoning. In the reasoning process, you can include detailed analysis, brainstorming, verification, and refinement of ideas.
You must put the final answer in \textbackslash boxed\{...\} on the last line.
Important: Do not put anything after the boxed answer. Let's think step by step.
\end{systempromptblock}

\begin{systempromptblock}{System prompt for evaluation}
Please reason step by step, and put your final answer within \textbackslash boxed\{...\}.
\end{systempromptblock}

\subsection{Detailed Experimental Setup and Hyperparameter Selection}
\paragraph{Reward function setting.}We adopt the most common reward setting in RLVR training, which is a 0-1 binary reward. The model's output answer is extracted using the Math-Verify tool and compared with the correct answer. If the two are exactly the same, give the model a reward of 1; if they are not the same, give the model a reward of 0. In addition, we also introduced a weak format reward. If the model can output \textbackslash boxed\{...\} at the end, give the model a format reward value of 0.05.

\paragraph{Hardware configuration.}For the Qwen2.5-Math-1.5B model, all experiments were conducted on 4 NVIDIA H100 GPUs with 80G of memory; for the Qwen3-4B-Base and Llama3.1-8B models, all experiments were carried out on 8 NVIDIA A100 GPUs with 40G of memory. All experiments were carried out on a single node. The training of the same series of models was conducted under the same computing resources and using exactly the same data preprocessing and evaluation standards to ensure the fairness of the comparison. 

\paragraph{Evaluation configuration.}For evaluation, we set the temperature to 0.6, random seed to 42, top-$p$ to 0.7, top-$k$ to -1 and the maximum number of new tokens to 8192. We implement the evaluation pipeline using vLLM for generation and math-verify for answer verification.

\paragraph{Evaluation Metrics for Maj and Avg.}
For each problem, we sample $K$ responses from the model under the same decoding configuration. Let $N$ denote the number of test problems and let $c_{i,j}\in\{0,1\}$ indicate whether the $j$-th sampled response for the $i$-th problem is correct after answer extraction and automatic verification. The \textit{Avg@K} metric measures the average correctness over all sampled responses:
\[
\mathrm{Avg@}K = \frac{1}{N}\sum_{i=1}^{N}\frac{1}{K}\sum_{j=1}^{K} c_{i,j}.
\]

For \textit{Maj@K}, we first extract the final answer from each of the $K$ sampled responses for a given problem and then apply majority voting over the extracted answers. 
The answer with the highest frequency is selected as the final prediction for that problem and is subsequently compared with the ground-truth answer. 
Formally, let $\hat{a}_{i,j}$ be the extracted answer from the $j$-th response to the $i$-th problem, and let $a_i^\star$ be the corresponding ground-truth answer. 
The majority-voted prediction is defined as
\[
\hat{a}^{\mathrm{maj}}_i
=
\arg\max_{a}
\sum_{j=1}^{K}
\mathbb{I}(\hat{a}_{i,j}=a),
\]
and \textit{Maj@K} is computed as
\[
\mathrm{Maj@}K
=
\frac{1}{N}
\sum_{i=1}^{N}
\mathbb{I}\left(
\hat{a}^{\mathrm{maj}}_i = a_i^\star
\right).
\]
When multiple answers receive the same number of votes, we use a deterministic tie-breaking rule based on their first occurrence among the sampled responses. 
Responses from which no valid final answer can be extracted are treated as incorrect for \textit{Avg@K} and are excluded from the majority candidates unless all sampled responses are invalid, in which case the problem is counted as incorrect for \textit{Maj@K}.

\paragraph{Supplementary Pass@K Evaluation.}
We additionally report Pass@K to measure sampling-dependent solution coverage. For each problem, Pass@K counts the prediction as correct if at least one of the $K$ sampled responses is verified as correct. This metric complements Maj@K and Avg@K by evaluating whether a model preserves access to correct reasoning trajectories under repeated sampling.

\begin{figure*}[t]
  \centering
  \captionsetup[subfigure]{labelformat=simple,labelsep=space}
  \renewcommand\thesubfigure{(\alph{subfigure})}
  \begin{subfigure}[t]{0.48\textwidth}
    \centering
    \includegraphics[width=\linewidth]{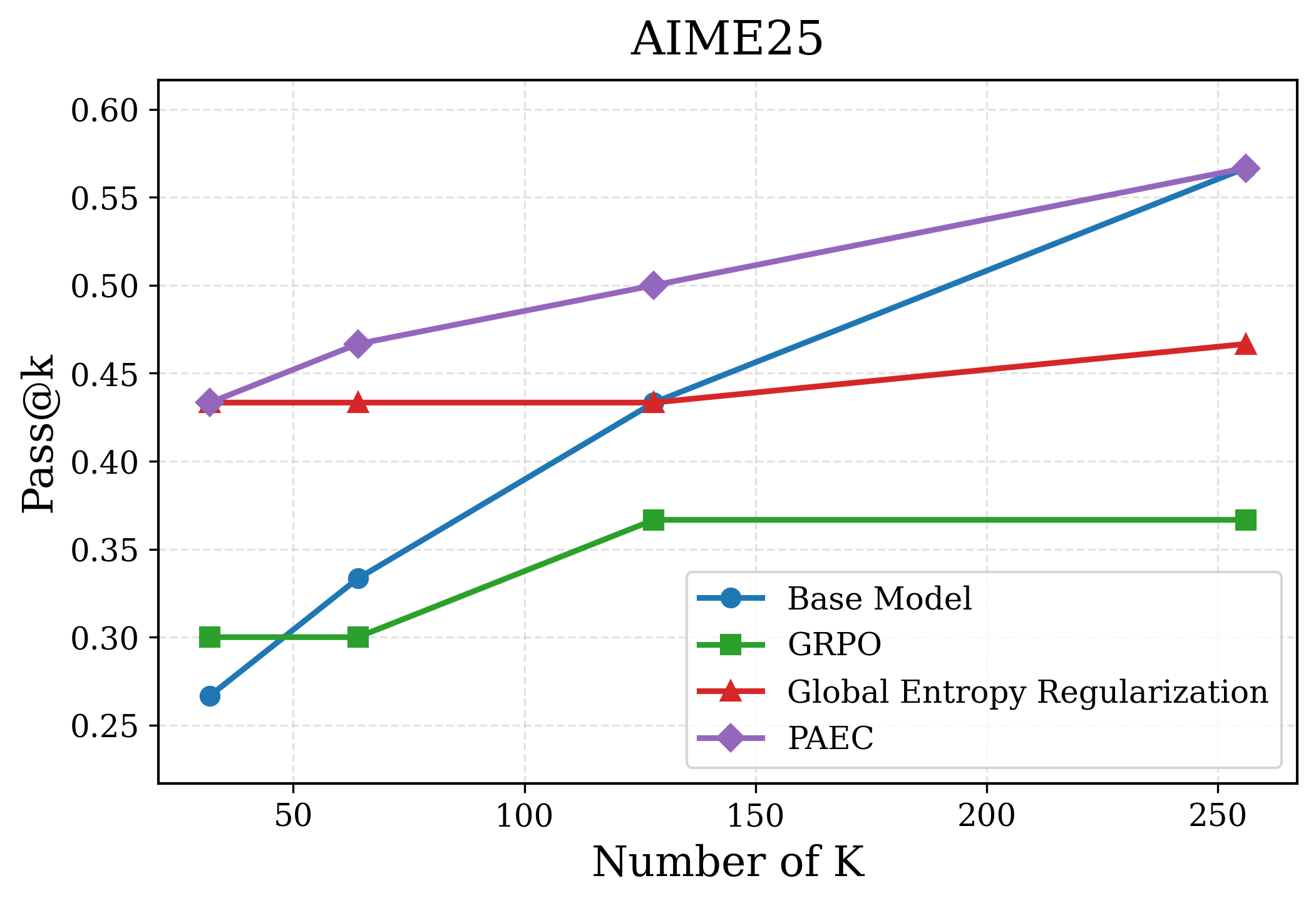}
    \caption{AIME25.}
    \label{fig:appendix_pass_k_aime25}
  \end{subfigure}
  \hfill
  \begin{subfigure}[t]{0.48\textwidth}
    \centering
    \includegraphics[width=\linewidth]{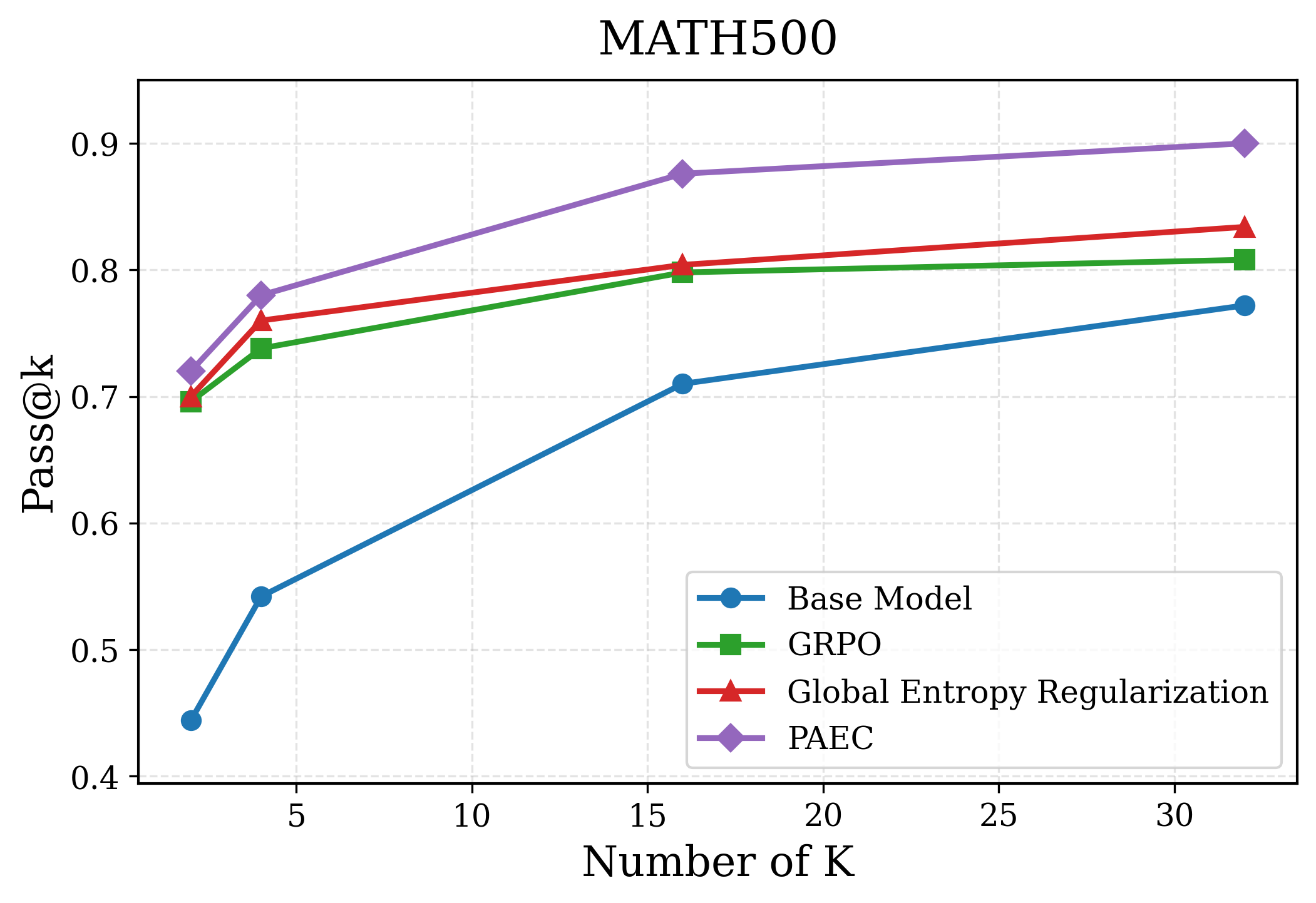}
    \caption{MATH500.}
    \label{fig:appendix_pass_k_math500}
  \end{subfigure}
  \caption{Supplementary Pass@K results on AIME25 and MATH500. PAEC achieves consistently strong sampling-based coverage, indicating that its improvements are not limited to majority voting.}
  \label{fig:appendix_pass_k}
\end{figure*}

\paragraph{Hyperparameter Settings for Baseline Methods.}
For a fair comparison, all baseline methods use the same training configuration as GRPO unless otherwise specified, including the model initialization, training data, rollout settings, optimizer, batch size, and evaluation protocol. We only modify the method-specific hyperparameters required by each baseline.

\textbf{CISPO.}
For CISPO, we set the asymmetric clipping thresholds to $\epsilon_{\mathrm{high}}=5.0$ and $\epsilon_{\mathrm{low}}=0.0$. All other hyperparameters are kept identical to those used in GRPO.

\textbf{Global Entropy Regularization.}
For the global entropy regularization baseline, we add a standard entropy bonus to the GRPO objective and set the entropy coefficient to $8\times10^{-4}$. We use this relatively small coefficient because overly strong global entropy regularization can induce excessive policy entropy and destabilize RLVR training, which may further degrade downstream reasoning performance. Except for the entropy coefficient, all other settings are the same as GRPO.

\textbf{Clip-Cov.}
For Clip-Cov, we follow the hyperparameter configuration recommended in the original paper and set $\texttt{clip\_cov\_ratio}=2\times10^{-4}$, $\texttt{clip\_cov\_lb}=1.0$, and $\texttt{clip\_cov\_ub}=5.0$. The remaining training and evaluation settings are kept consistent with GRPO.

\textbf{KL-Cov.}
For KL-Cov, we also adopt the recommended hyperparameters from the original paper, with $\texttt{kl\_cov\_ratio}=2\times10^{-4}$ and $\texttt{ppo\_kl\_coef}=0.1$. All other hyperparameters remain unchanged from the GRPO configuration.

\textbf{AER.}
For AER, we reproduce the method within the VeRL framework according to the algorithmic description in the original paper. We use the default hyperparameters recommended by the authors, setting $\tau=0.4$, $\rho=0.2$, and $\eta=0.005$. All other experimental settings are aligned with GRPO to ensure a controlled comparison.

\paragraph{Configuration of Qwen series model experiment.}For our experiments, we use the VeRL reinforcement learning framework, which is an open source and practical framework for large language model post-training. It integrates many reinforcement learning post-training algorithms. We use GRPO with both clipping thresholds set to 0.2, AdamW with a learning rate of $1 \times 10^{-6}$, a total batch size of 96, and a PPO minibatch size of 24. For each prompt, we sample 8 responses during training. The maximum prompt length and response length are set to 512 and 3072. GRPO rollout top-$p$ is set to 1.0 and top-$k$ is set to -1. GRPO rollout temperature is set to 1.0. All models in this setting are trained for 720 update steps. We save the checkpoint every 10 training steps and evaluate the model on the validation set (AIME23 is used as the validation set to better reflect the performance of the model on problems that are close to the benchmark difficulty. Because the dataset does not provide LaTeX source references, we provide the source link here\footnote{\url{https://huggingface.co/datasets/extraordinarylab/aime23}}). The validation sampling temperature, top-$p$, top-$k$ and maximum number of new tokens are set to be consistent with the overall evaluation, and their Pass@16 indicators are monitored. In the final evaluation, the best performing checkpoint on the validation set is selected.

\paragraph{Other Hyperparameter Considerations.}
Table~\ref{tab:robustness_full} reports the complete performance of the model for different values of two of the more critical and sensitive hyperparameters introduced in the main text.
\begin{table*}[t]
\centering
\setlength{\tabcolsep}{4pt}
\renewcommand{\arraystretch}{1.2}
\resizebox{\textwidth}{!}{%
\begin{tabular}{l|cc|cc|cc|cc|cc|cc}
\hline
\multirow{2}{*}{Setting} & \multicolumn{2}{c|}{AIME24} & \multicolumn{2}{c|}{AIME25} & \multicolumn{2}{c|}{AIME26} & \multicolumn{2}{c|}{MATH500} & \multicolumn{2}{c|}{AMC23} & \multicolumn{2}{c}{Average} \\
 & Maj@32 & Avg@32 & Maj@32 & Avg@32 & Maj@32 & Avg@32 & Maj@8 & Avg@8 & Maj@16 & Avg@16 & Maj & Avg \\
\hline
PAEC($\beta$=0.3) & 26.7 & 13.6 & 26.7 & 11.8 & 16.7 & 9.3 & 75.6 & 71.0 & 62.5 & 58.0 & \textbf{41.6} & \textbf{32.7} \\
PAEC($\beta$=0.7) & 26.7 & 15.0 & 23.3 & 9.9 & 13.3 & 8.4 & 73.4 & 69.6 & 60.0 & 57.3 & 39.3 & 32.0 \\
PAEC($\beta$=0.0) & 23.3 & 16.3 & 20.0 & 8.9 & 10.0 & 8.1 & 74.2 & 68.2 & 62.5 & 56.6 & 38.0 & 31.6 \\
\hline
PAEC(quantile=0.8) & 26.7 & 13.6 & 26.7 & 11.8 & 16.7 & 9.3 & 75.6 & 71.0 & 62.5 & 58.0 & \textbf{41.6} & \textbf{32.7} \\
PAEC(quantile=0.3) & 20.0 & 13.4 & 16.7 & 9.2 & 13.3 & 8.8 & 70.4 & 65.5 & 62.5 & 55.9 & 36.6 & 30.6 \\
\hline
\end{tabular}%
}
\caption{Complete results of the hyperparameter robustness analysis for PAEC under different competition weights $\beta$ and mask quantile thresholds.}
\label{tab:robustness_full}
\end{table*}

Table~\ref{tab:paec_hyperparams} reports the default auxiliary hyperparameters used in PAEC. Since an exhaustive robustness analysis over all auxiliary hyperparameters would introduce substantial additional computational cost, we keep these values fixed across all main experiments unless otherwise specified. 
\begin{table}[htbp]
\centering
\small
\begin{tabular}{lcc}
\toprule
Hyperparameter & Value & Symbol \\
\midrule
\texttt{peak\_prob\_p} & 0.85 & $p_{\mathrm{peak}}$ \\
\texttt{peak\_prob\_topk} & 10000 & $k_{\mathrm{peak}}$ \\
\texttt{soft\_mask\_sharpness} & 12.0 & $\kappa$ \\
\texttt{paec\_alpha} & 0.0008 & $\alpha$ \\
\texttt{paec\_lambda\_ent} & 0.004 & $\lambda_{\mathrm{ent}}$ \\
\texttt{paec\_h0\_init\_steps} & 4 & $K$ \\
\bottomrule
\end{tabular}
\caption{Default auxiliary hyperparameters used in PAEC.}
\label{tab:paec_hyperparams}
\end{table}

The hyperparameters \texttt{peak\_prob\_p} and \texttt{peak\_prob\_topk} define the high-probability token support on which the local entropy is computed. We set \texttt{peak\_prob\_p} to $0.85$, which preserves most of the meaningful probability mass while excluding the long-tail region of the vocabulary. In practice, a substantially smaller value may over-truncate the candidate set and underestimate the uncertainty of the policy, whereas a value close to $1.0$ makes the computation approach full-vocabulary entropy and weakens the intended focus on the effective action space. 
The parameter \texttt{peak\_prob\_topk} is set to $10000$ as a conservative upper bound. 

The parameter \texttt{soft\_mask\_sharpness} controls the steepness of the sigmoid function used to construct the position-aware soft mask. A smaller value produces a flatter mask and makes the weighting closer to a uniform entropy regularizer, thereby reducing the selectivity of PAEC over critical token positions. In contrast, an excessively large value makes the mask close to a hard threshold, which may increase sensitivity to small fluctuations in the token-level scores. We therefore use $\kappa=12.0$ as a moderately sharp setting, which provides sufficient contrast between critical and non-critical positions while preserving the smoothness of soft masking.

The coefficient \texttt{paec\_alpha} determines the strength of the selective entropy regularization term. 
We set $\alpha=8\times10^{-4}$ so that the entropy term can encourage exploration at critical positions without dominating the original policy optimization objective. 
A much smaller coefficient may make the entropy regularizer ineffective, while a much larger coefficient may over-emphasize exploration and degrade answer accuracy.

The coefficient \texttt{paec\_lambda\_ent} controls the one-sided entropy lower-bound penalty. This term is activated only when the aggregated masked entropy falls below the initial entropy anchor. We use $\lambda_{\mathrm{ent}}=0.004$ to provide a mild but effective constraint on critical-position entropy. If this coefficient is too small, the controller may fail to prevent entropy collapse; if it is too large, the penalty may dominate the policy-gradient update and destabilize training.

Finally, \texttt{paec\_h0\_init\_steps} specifies the number of initial training steps used to estimate the entropy anchor $H_0$. We set $K=4$ in our experiments. The purpose of this short warm-up period is to estimate the initial entropy level before the policy distribution undergoes substantial drift. Using too few steps may lead to a noisy anchor estimate, while using too many steps may incorporate entropy values from an already updated policy and thus weaken the role of $H_0$ as an initial reference. Setting this value too large is not recommended.

\paragraph{Effect of Response Length Truncation.}
Due to computational constraints, we set the maximum response length during rollout to 3072 tokens in our main experiments. 
This limit is shorter than the commonly used long-context generation budget, such as 8192 tokens, and may therefore truncate a non-negligible fraction of long reasoning trajectories, especially at the early stage of training. When a response is truncated before reaching a final answer, the verifier cannot reliably assign a positive reward even if the partial reasoning trajectory is potentially useful. As a result, the response-length constraint may introduce an implicit selection pressure against excessively long generations.

\begin{figure*}[t]
  \centering
  \captionsetup[subfigure]{labelformat=simple,labelsep=space}
  \renewcommand\thesubfigure{(\alph{subfigure})}
  \begin{subfigure}[t]{0.48\textwidth}
    \centering
    \includegraphics[width=\linewidth]{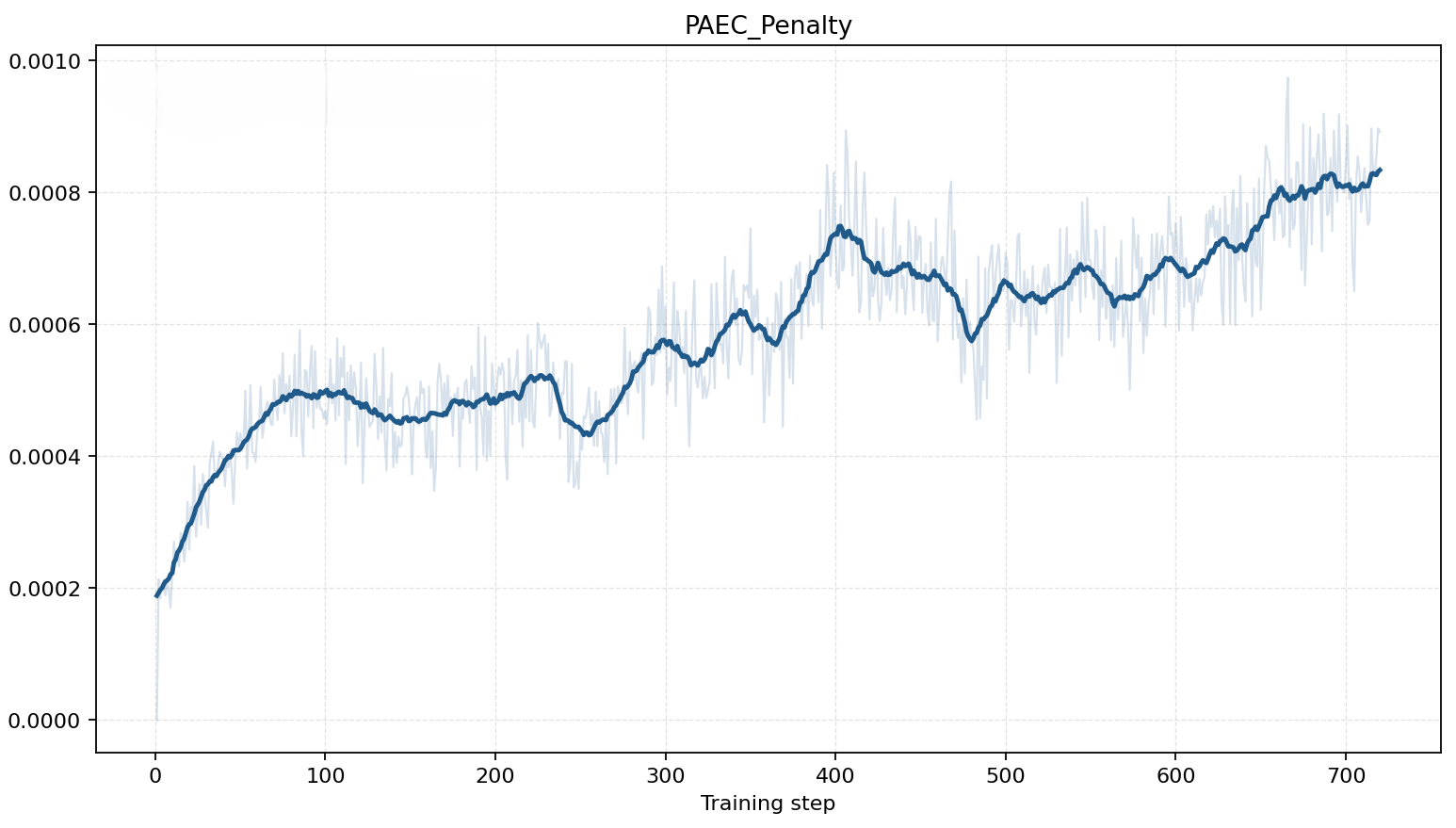}
    \caption{PAEC penalty.}
    \label{fig:appendix_paec_penalty}
  \end{subfigure}
  \hfill
  \begin{subfigure}[t]{0.48\textwidth}
    \centering
    \includegraphics[width=\linewidth]{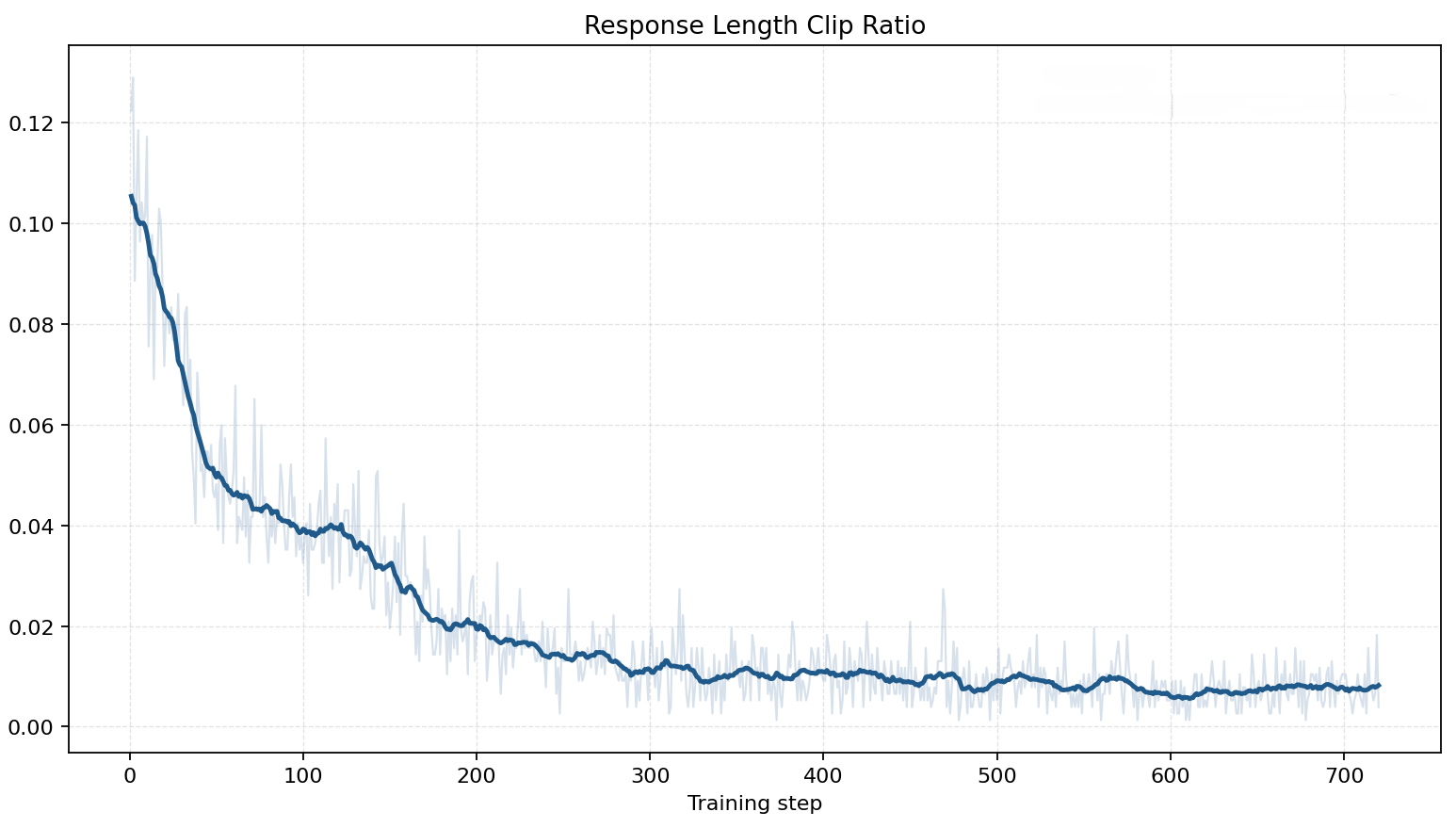}
    \caption{Response length clip ratio.}
    \label{fig:appendix_clip_ratio}
  \end{subfigure}

  \vspace{0.8em}

  \begin{subfigure}[t]{0.48\textwidth}
    \centering
    \includegraphics[width=\linewidth]{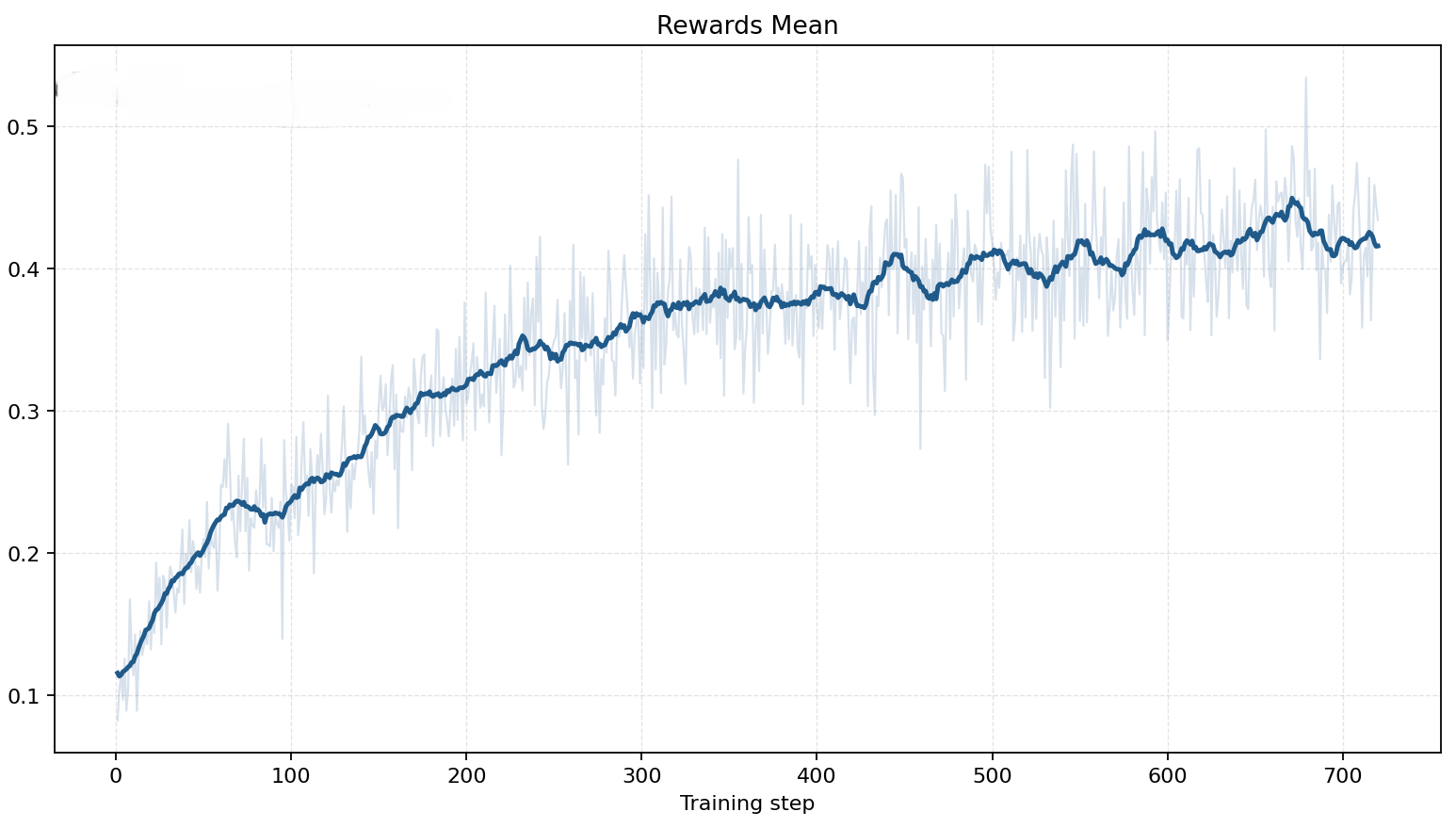}
    \caption{Mean reward.}
    \label{fig:appendix_rewards_mean}
  \end{subfigure}
  \hfill
  \begin{subfigure}[t]{0.48\textwidth}
    \centering
    \includegraphics[width=\linewidth]{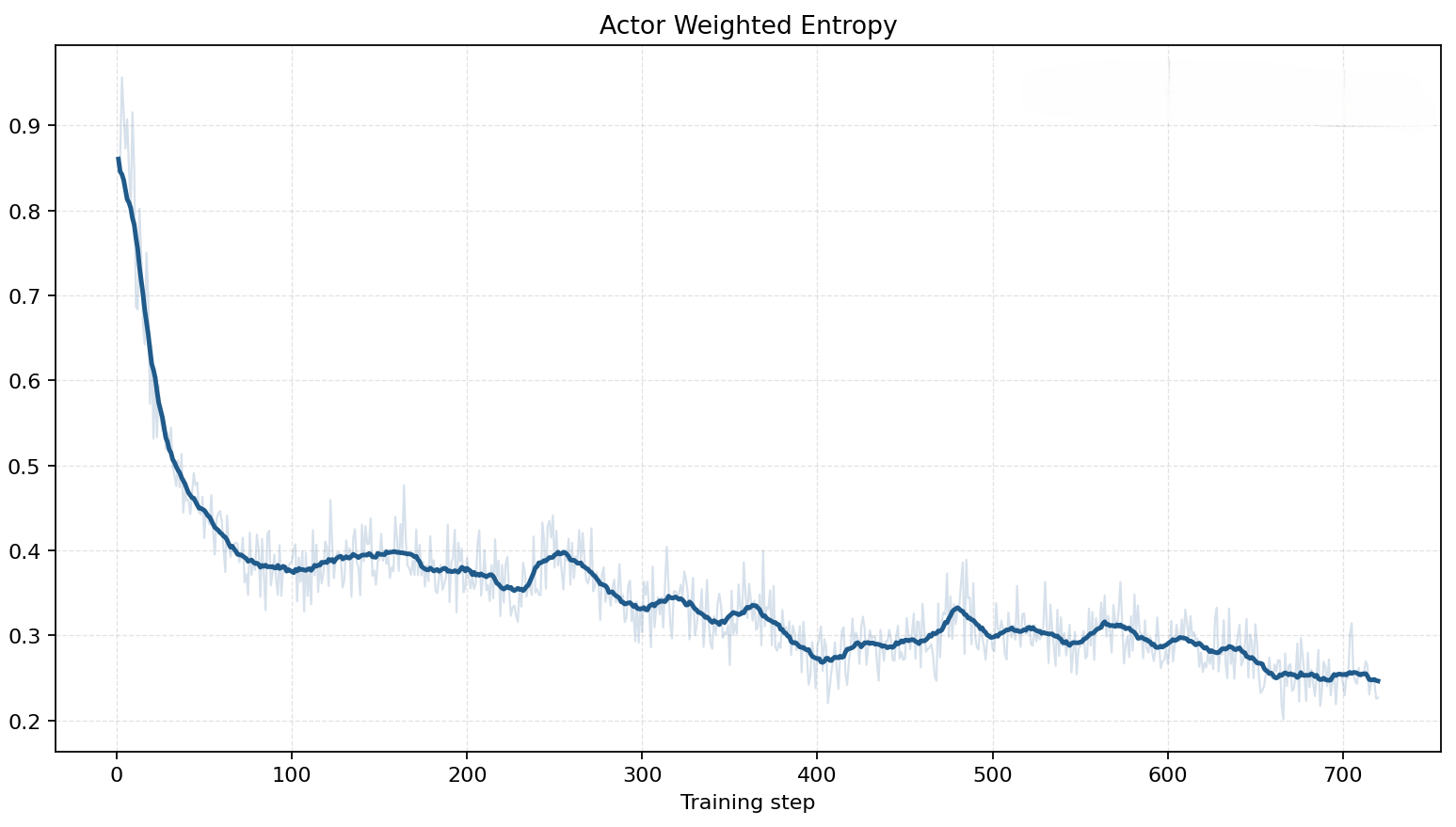}
    \caption{Actor weighted entropy.}
    \label{fig:appendix_weighted_entropy}
  \end{subfigure}
  \caption{Training dynamics of PAEC on Qwen2.5-Math-1.5B. The four panels show the PAEC lower-bound penalty, response length clip ratio, mean reward, and actor weighted entropy during RLVR training.}
  \label{fig:appendix_training_dynamics}
\end{figure*}

To better understand this effect, we report the response truncation ratio throughout training in Figure~\ref{fig:appendix_training_dynamics}(b). The curve shows that the truncation ratio is relatively high at the beginning of training, with approximately one-tenth of the responses being truncated, indicating that the initial policy produces responses that exceed the rollout length budget. As training proceeds, the truncation ratio decreases rapidly and remains at a low level in the later stage. This trend suggests that the trained policy gradually adapts to the available generation budget and produces fewer over-length responses. Importantly, the truncation ratio does not increase in the later stage, indicating that the observed performance improvements are unlikely to be driven by uncontrolled growth in response length. Nevertheless, we acknowledge that a larger response-length budget may further benefit models that rely on longer reasoning chains, and leave a systematic study of longer rollout contexts to future work.

\begin{figure*}[t]
  \centering
  \captionsetup[subfigure]{labelformat=simple,labelsep=space}
  \renewcommand\thesubfigure{(\alph{subfigure})}
  \begin{subfigure}[t]{0.48\textwidth}
    \centering
    \includegraphics[width=\linewidth]{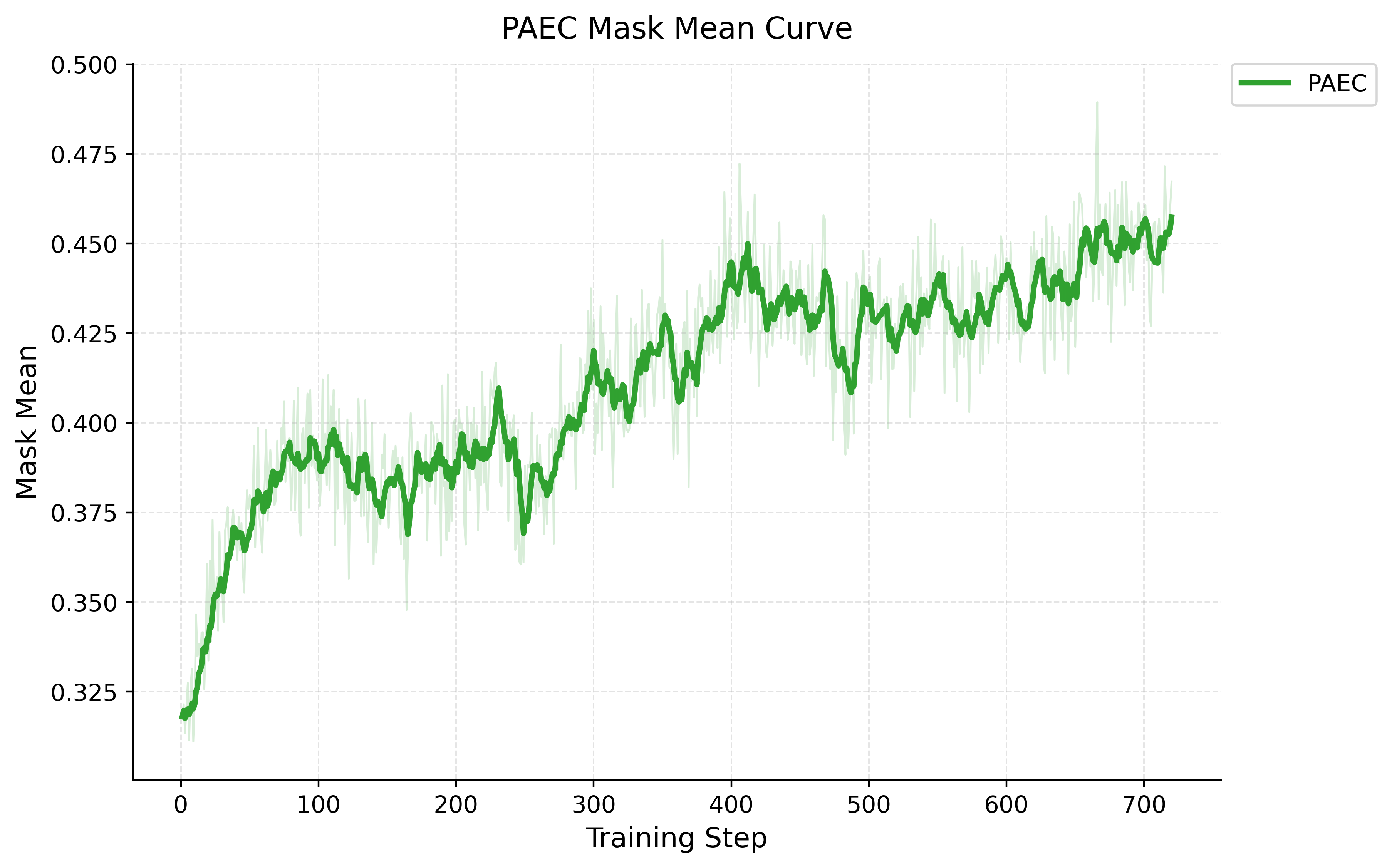}
    \caption{Mean soft-mask value during PAEC training.}
    \label{fig:appendix_mask_curve}
  \end{subfigure}
  \hfill
  \begin{subfigure}[t]{0.48\textwidth}
    \centering
    \includegraphics[width=\linewidth]{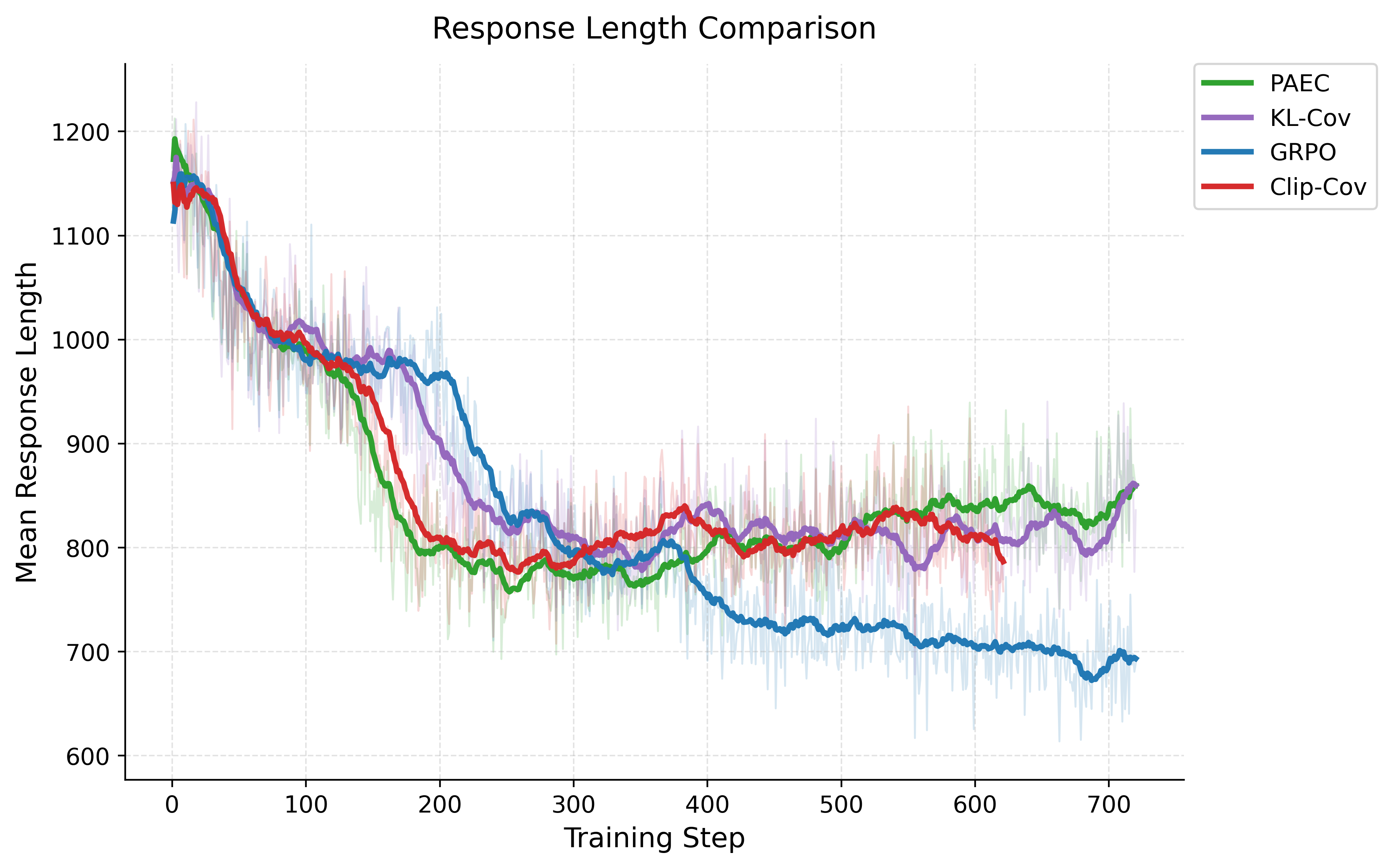}
    \caption{Average response length of different RLVR methods.}
    \label{fig:appendix_response_curve}
  \end{subfigure}
  \caption{Additional training diagnostics. The mask-mean curve shows that PAEC maintains continuous and non-degenerate position weights, while the response-length curve shows that entropy-preserving methods exhibit mild later-stage length recovery under the same rollout budget.}
  \label{fig:appendix_mask_response}
\end{figure*}

\paragraph{Additional Training Diagnostics.}
In addition to the truncation ratio, we report two complementary training diagnostics in Figure~\ref{fig:appendix_mask_response}. The mask-mean curve tracks how strongly PAEC weights selected positions during training, while the response-length curve compares the generation-length dynamics of different RLVR methods. These diagnostics help interpret the main results without treating response length itself as direct evidence of reasoning quality.

\subsection{Supplementary Experiments on Different Model Scales}
We conduct supplementary experiments on Qwen3-4B-Base and Llama3.1-8B-Base. Due to computational constraints, we focus on three representative methods in this supplementary study: GRPO, Global Entropy Regularization, and PAEC. All methods are trained and evaluated under the same protocol within each model family, using the same data preprocessing, rollout configuration, training budget, and evaluation settings.

\paragraph{Configuration of Llama series model experiment.}
We use the VeRL reinforcement learning framework. We use GRPO with both clipping thresholds set to 0.2, AdamW with a learning rate of $1 \times 10^{-6}$, a total batch size of 36, and a PPO minibatch size of 6. For each prompt, we sample 8 responses during training. The maximum prompt length and response length are set to 512 and 2048. GRPO rollout top-$p$ is set to 1.0 and top-$k$ is set to -1. GRPO rollout temperature is set to 1.0. Respectively, all models are trained for 1100 update steps. We save the checkpoint every 10 training steps and evaluate the model on the validation set. The validation sampling temperature, top-$p$, top-$k$ and maximum number of new tokens are set to be consistent with the overall evaluation, and their Pass@16 indicators are monitored. In the final evaluation, the best performing checkpoint on the validation set is selected.

Table~\ref{tab:overall_performance} reports the complete results. 

\begin{table*}[t]
\centering
\small
\setlength{\tabcolsep}{4.2pt}
\renewcommand{\arraystretch}{1.08}
\resizebox{\textwidth}{!}{
\begin{tabular}{lcccccccccccc}
\toprule
\multirow{2}{*}{\textbf{Method}}
& \multicolumn{2}{c}{\textbf{AIME24}}
& \multicolumn{2}{c}{\textbf{AIME25}}
& \multicolumn{2}{c}{\textbf{AIME26}}
& \multicolumn{2}{c}{\textbf{MATH500}}
& \multicolumn{2}{c}{\textbf{AMC23}}
& \multicolumn{2}{c}{\textbf{Avg.}} \\
\cmidrule(lr){2-3}
\cmidrule(lr){4-5}
\cmidrule(lr){6-7}
\cmidrule(lr){8-9}
\cmidrule(lr){10-11}
\cmidrule(lr){12-13}
& \textit{Maj@32} & \textit{Avg@32}
& \textit{Maj@32} & \textit{Avg@32}
& \textit{Maj@32} & \textit{Avg@32}
& \textit{Maj@8} & \textit{Avg@8}
& \textit{Maj@16} & \textit{Avg@16}
& \textit{Maj} & \textit{Avg} \\
\midrule

\multicolumn{13}{c}{\textit{Qwen3-4B-Base}} \\
\midrule
Base
& 56.7 & 46.5
& 50.0 & 35.9
& 50.0 & 38.4
& 86.8 & 84.2
& 90.0 & 82.9 
& 66.7 & 57.6 \\

GRPO
& 60.0 & 46.5
& 53.3 & 39.2
& 63.3 & 42.9
& 88.2 & 86.4
& 95.0 & 87.2
& 71.9 & 60.4 \\

Global Entropy Reg
& 60.0 & 46.9
& 56.7 & 40.8
& 53.3 & 40.3
& 88.6 & 86.4
& 90.0 & 82.9
& 69.7 & 59.5 \\

\rowcolor{gray!15}
PAEC (Ours)
& 66.7 & 48.3
& 60.0 & 41.9
& 66.7 & 43.2
& 90.0 & 86.8
& 92.5 & 87.8
& 75.2 & 62.0 \\

\midrule
\multicolumn{13}{c}{\textit{Llama3.1-8B}} \\
\midrule
Base
& 0.0 & 0.0
& 3.3 & 0.4
& 0.0 & 0.1
& 9.8 & 4.6
& 5.0 & 3.0
& 3.6 & 1.6 \\

GRPO
& 3.3 & 0.2
& 0.0 & 0.0
& 3.3 & 0.2
& 25.6 & 21.3 
& 7.5 & 7.2
& 7.9 & 5.8 \\

Global Entropy Reg
& 3.3 & 0.2
& 0.0 & 0.1
& 3.3 & 0.5
& 20.6 & 18.4
& 10.0 & 8.4
& 7.4 & 5.5 \\

\rowcolor{gray!15}
PAEC (Ours)
& 3.3 & 0.1
& 3.3 & 0.0
& 6.7 & 1.0
& 26.4 & 21.8
& 15.0 & 10.4
& 10.9 & 6.7 \\

\bottomrule
\end{tabular}
}
\caption{\textbf{Overall performance comparison.}
The complete experimental results comparison of the three methods, namely GRPO, Global Entropy Regularization and PAEC, on the Qwen3-4B-Base and Llama3.1-8B models.}
\label{tab:overall_performance}
\end{table*}
On Qwen3-4B-Base, PAEC achieves the best macro-average performance, obtaining $75.2$ Maj and $62.0$ Avg. 
Compared with GRPO, PAEC improves the macro-average Maj score by $3.3$ points and the macro-average Avg score by $1.6$ points. Compared with Global Entropy Regularization, PAEC improves macro-average Maj by $5.5$ points and macro-average Avg by $2.5$ points. The improvement is especially clear on AIME-style benchmarks, where PAEC obtains higher Maj scores on AIME24, AIME25, and AIME26. These results suggest that position-aware entropy allocation remains beneficial when moving from the 1.5B-scale mathematical model to a stronger 4B-scale backbone.

On Llama3.1-8B-Base, the absolute mathematical reasoning performance is much lower than that of Qwen-series mathematical models. Nevertheless, PAEC still achieves the best macro-average performance among the compared methods, reaching $10.9$ Maj and $6.7$ Avg. In comparison, GRPO obtains $7.9$ Maj and $5.8$ Avg, while Global Entropy Regularization obtains $7.4$ Maj and $5.5$ Avg. Although the absolute scores are limited, the relative trend is consistent with the main experiments: selective, position-aware entropy calibration provides a more effective exploration mechanism than uniform entropy regularization.

Overall, these supplementary experiments indicate that the benefit of PAEC is not restricted to a single backbone. 
However, we emphasize that these results should be interpreted as supporting evidence rather than a complete scaling study, because the supplementary experiments cover fewer baselines and are conducted under limited computational budgets. A more systematic analysis over larger models and longer training schedules is left for future work.

\clearpage
\onecolumn
\nolinenumbers
\begin{multicols*}{2}
\raggedcolumns
\section{Case Study}
\newtcolorbox{Response}[1]{%
    enhanced,
    colback=green!10,
    colbacktitle=green!60!black,
    coltitle=white,
    colframe=green!60!black,
    fonttitle=\bfseries\sffamily,
    title={Response from Qwen2.5-Math-1.5B \hfill #1},
    arc=2pt,
    boxrule=0.5pt,
    breakable,
    before skip=4pt,
    after skip=4pt,
    before upper={\setlength{\parskip}{0pt}\sloppy},
    width=.96\linewidth,
    center,
}
% 2. 问题/案例区块
\newtcolorbox{QuestionBlock}{%
    enhanced,
    colback=gray!5,
    colframe=gray!30,
    coltitle=black,
    fonttitle=\bfseries,
    title={Question},
    arc=2pt,
    boxrule=0.5pt,
    breakable,
    before skip=4pt,
    after skip=4pt,
    before upper={\sloppy},
}

% ====================== 具体内容 ======================
To provide a qualitative example of the reasoning behavior produced by PAEC, we present one representative response in this section. The goal of this case study is not to provide causal evidence for the mask assignment, but to illustrate that PAEC can produce a complete and concise reasoning trajectory under the rollout length constraint.

\paragraph{Case 1: The correct example.}
\vspace{-0.5em}
\begin{QuestionBlock}
Find the sum of all integer bases $b>9$ for which $17_b$ is a divisor of $97_b.$
\textbf{Ground Truth:} 70
\end{QuestionBlock}
\begin{Response}{Total Length: 776 Tokens}
To solve \heavytoken{the} problem, we need to \heavytoken{convert} the numbers $17_b$ and $97_b$ from base $b$ to base 10 \heavytoken{and} then \heavytoken{determine the} values of $b$ for which $17_b$ divides $97_b$.

1. \heavytoken{Convert} $17_b$ to \heavytoken{its} base 10 equivalent:
\[ 17_b = 1 \cdot b + 7 = b + 7. \]

2. Convert $97_b$ to its base 10 equivalent:
\[ 97_b = 9 \cdot b + 7 = 9b + 7. \]

3. \heavytoken{We} need \heavytoken{to} find the values of $b$ \heavytoken{for} which $b + 7$ divides $9b + 7$. This means that there exists an integer $k$ such that:
\[ 9b + 7 = k(b + 7). \]
Rearranging \heavytoken{this} equation, we get:
\[ 9b + 7 = kb + 7k, \]
\[ 9b - kb = 7k - 7, \]
\[ b(9 - k) = 7(k - 1). \]
\heavytoken{Therefore}:
\[ b = \frac{7(k - 1)}{9 - k}. \]
For $b$ to be an integer, $\frac{7(k - 1)}{9 - k}$ must be an integer. This means that $9 - k$ must be a divisor of $7(k - 1)$. \heavytoken{Let}'s find the \mediumtoken{possible} values of $k$.

\heavytoken{Since} $b > 9$, we have:
\[ \frac{7(k - 1)}{9 - k} > 9. \]
This \heavytoken{implies}:
\[ 7(k - 1) > 9(9 - k), \]
\[ 7k - 7 > 81 - 9k, \]
\[ 16k > 88, \]
\[ k > 5.5. \]
\heavytoken{Since} $k$ is an integer, $k \geq 6$. \heavytoken{Also}, \heavytoken{since} $9 - k$ must \heavytoken{be} a divisor of $7(k - 1)$, \heavytoken{let}'s check the \mediumtoken{possible} values of $k$ starting from 6.

- If $k = 6$, then $b = \frac{7(6 - 1)}{9 - 6} = \frac{35}{3}$, which is not an integer.
- If $k = 7$, then $b = \frac{7(7 - 1)}{9 - 7} = \frac{42}{2} = 21$, which is an integer and greater than 9.
- If $k = 8$, then $b = \frac{7(8 - 1)}{9 - 8} = \frac{49}{1} = 49$, which is an integer and \heavytoken{greater} than 9.
- If $k = 9$, then $b = \frac{7(9 - 1)}{9 - 9}$, which is \mediumtoken{undefined}.
- If $k = 10$, then $b = \frac{7(10 - 1)}{9 - 10} = \frac{63}{-1} = -63$, which is not a valid base.

So the \heavytoken{valid} values of $k$ are 7 and 8, giving us \heavytoken{the corresponding} values of $b$ as 21 and 49. The sum of these values is:
\[ 21 + 49 = 70. \]

\heavytoken{Therefore}, the final answer is:
\boxed{70}
\end{Response}

\paragraph{Case 2: The incorrect example.}
\vspace{-0.5em}
\begin{QuestionBlock}
The $9$ members of a baseball team went to an ice-cream parlor after their game. Each player had a single scoop cone of chocolate, vanilla, or strawberry ice cream. At least one player chose each flavor, and the number of players who chose chocolate was greater than the number of players who chose vanilla, which was greater than the number of players who chose strawberry. Let $N$ be the number of different assignments of flavors to players that meet these conditions. Find the remainder when $N$ is divided by $1000.$
\textbf{Ground Truth:} 16
\end{QuestionBlock}
\begin{Response}{Total Length: 834 Tokens}
\mediumtoken{To} solve \mediumtoken{this} problem, we need to \heavytoken{determine} the number of \mediumtoken{ways} to assign 9 players to 3 flavors (chocolate, vanilla, and strawberry) such that \heavytoken{each} flavor is chosen \mediumtoken{by} at least one player, and the number of players \mediumtoken{choosing chocolate} is greater than the number of players choosing vanilla, which \heavytoken{in} turn is greater than the number of players choosing strawberry. \heavytoken{Let's} \mediumtoken{denote} the number of players choosing chocolate, vanilla, and strawberry as \(c\), \(v\), and \(s\) respectively. We \heavytoken{need} to \heavytoken{find the} number of \heavytoken{solutions} to the equation \(c + v + s = 9\) with the constraints \(c > v > s \geq 1\).

\heavytoken{First}, \heavytoken{we will enumerate} all possible values for \(c\), \(v\), and \(s\) that satisfy \heavytoken{these constraints}. Then, \heavytoken{for} each valid combination, we will calculate the number of ways to assign the players to these flavors. The number of ways to assign \(c\) players to chocolate, \(v\) players to vanilla, and \(s\) players to strawberry is given by the multinomial coefficient \(\frac{9!}{c!v!s!}\).

\heavytoken{Let}'s \heavytoken{enumerate} \mediumtoken{the possible} values for \(c\), \(v\), and \(s\):

1. \(c = 5\), \(v = 3\), \(s = 1\)

2. \(c = \heavytoken{6}\), \(v = 2\), \(s = 1\)

3. \(c = 7\), \(v = 2\), \(s = 0\) (\heavytoken{not} valid \mediumtoken{since} \(s \geq 1\))

4. \(c = \heavytoken{6}\), \(v = 3\), \(s = 0\) (not valid since \(s \geq 1\))

5. \(c = 5\), \(v = 4\), \(s = 0\) (not valid since \(s \geq 1\))

So, \heavytoken{we} \mediumtoken{only} \heavytoken{have} two valid combinations: \((c, v, s) = (5, 3, 1)\) and \((c, v, s) = (6, 2, 1)\).

\heavytoken{Next}, we \heavytoken{calculate} the number of ways to assign the players \heavytoken{for} each \heavytoken{combination}:

1. For \((c, v, s) = (5, 3, 1)\), the number of ways is \(\frac{9!}{5!3!1!} = 504\).

2. For \((c, v, s) = (6, 2, 1)\), the number of ways is \(\frac{9!}{6!2!1!} = 336\).

\heavytoken{So}, the total number of \heavytoken{different} assignments \heavytoken{is} \(504 + 336 = 840\). The remainder when 840 is divided by 1000 is 840.

Let's \heavytoken{implement} this in Python to \heavytoken{confirm} our \heavytoken{result}.

\begin{lstlisting}[language=Python]
import math
# \heavytoken{Calculate} the number of ways for each valid combination
ways_5_3_1 = math.factorial(9) // (math.factorial(5) * math.factorial(3) * math.factorial(1))
ways_6_2_1 = math.factorial(9) // (math.factorial(6) * math.factorial(2) * math.factorial(1))

# Total number of different assignments
total_ways = ways_5_3_1 + ways_6_2_1
# Remainder when divided by 1000
remainder = total_ways % 1000

print((total_ways, remainder))
\end{lstlisting}

\begin{verbatim}
    (840, 840)
\end{verbatim}
The total number of different assignments of flavors to players \mediumtoken{that} meet the \heavytoken{given} conditions is 840. The remainder when 840 is divided by 1000 is \(\boxed{840}\)
\end{Response}
In the visualization, darker red indicates larger mask values and thus stronger entropy regularization, while lighter red or uncolored tokens correspond to lower mask values.
\end{multicols*}

\end{document}